\def\ColorScheme{purple}

\documentclass[11pt,logo,letterpaper]{berkeley}

\usepackage[comma,numbers,sort]{natbib}
\usepackage{booktabs}       %
\usepackage{nicefrac}       %
\usepackage{wrapfig}

\usepackage{multirow}
\usepackage{colortbl}   
\usepackage{tabularx}
\usepackage{float}
\usepackage{placeins}
\usepackage{tcolorbox}
\usepackage{makecell}
\usepackage{xspace}

\usepackage{fontawesome5}
\usepackage{longtable}

\usepackage{tcolorbox}
\tcbuselibrary{breakable, skins} 
\usepackage{listings}
\lstdefinestyle{prompt}{
    breaklines=true,
    breakatwhitespace=false,
    basicstyle=\tiny\ttfamily,
    frame=none,
    columns=flexible,
    breakindent=0pt,
    literate={—}{{-}}1
}
\makeatletter
\newcommand*\myfontsize{%
  \@setfontsize\myfontsize{7}{8}%
}
\makeatother

\definecolor{blanchedalmond}{rgb}{1.0, 0.92, 0.8}
\definecolor{carmine}{rgb}{0.59, 0.0, 0.09}
\definecolor{lightblue}{rgb}{0.22,0.45,0.70}%

\renewcommand{\mathbf}{\boldsymbol}

\makeatletter
\def\Ddots{\mathinner{\mkern1mu\raise\p@
\vbox{\kern7\p@\hbox{.}}\mkern2mu
\raise4\p@\hbox{.}\mkern2mu\raise7\p@\hbox{.}\mkern1mu}}
\makeatother

\definecolor{amaranth}{rgb}{0.9, 0.17, 0.31}
\definecolor{antiquebrass}{rgb}{0.8, 0.58, 0.46}
\definecolor{antiquefuchsia}{rgb}{0.57, 0.36, 0.51}
\definecolor{chromeyellow}{rgb}{0.31, 0.47, 0.26}

\definecolor{darkgreenrank}{RGB}{0,100,0}
\definecolor{darkbluerank}{RGB}{0,70,140}
\definecolor{darkorangerank}{RGB}{180,90,0}
\definecolor{bestblue}{HTML}{E8F2FF}

\newcommand{\best}[1]{{\color{darkgreenrank}\textbf{\underline{#1}}}}
\newcommand{\second}[1]{{\color{darkbluerank}\underline{#1}}}
\newcommand{\third}[1]{{\color{darkorangerank}\underline{#1}}}

\title{Detect by Yourself: Self-Designing Agentic Workflows for Few-Shot Graph Anomaly Detection}
\runningtitle{Detect by Yourself: Self-Designing Agentic Workflows for Few-Shot Graph Anomaly Detection}
\author{
    Tairan Huang$^{1}$,
    Qiang Chen$^{2}$,
    Yili Wang$^{3}$,
    Yueyue Ma$^{1}$,
    Changlong He$^{1}$,
    Xiu Su$^{1,*}$,
    Yi Chen$^{2,*}$
}
\affil{$^1$Central South University \quad $^2$The Hong Kong University of Science and Technology \quad $^3$The Hong Kong University of Science and Technology (Guangzhou)}

\correspondingauthor{$^*$Corresponding Author}

\begin{document}

\begin{abstract}

Graph anomaly detection aims to identify anomaly nodes in attributed graphs and plays an important role in real-world applications. 
However, existing graph anomaly detection methods still face two key challenges: 
1) \textit{fixed pipelines}, which restrict their adaptability across different graph tasks under limited supervision; 
2) \textit{weak evidence}, which prevents them from explicitly incorporating contextual and structural anomaly signals into the detection process.
In this paper, we propose a novel framework, self-designing agentic workflows for few-shot graph anomaly detection (\textbf{SignGAD}). 
Specifically, we propose a novel paradigm that reformulates graph anomaly detection task from training a fixed anomaly detector to designing task-conditioned detection workflows.
By constructing detection workflows, SignGAD selects suitable graph encodings and detector designs to exploit task-specific anomaly evidence.
Meanwhile, we introduce a guarded final refit strategy to refine the selected workflow by calibrating refit acceptance, enhancing reliability under limited supervision.
Extensive experiments conducted on several real-world datasets demonstrate that SignGAD achieves strong performance against state-of-the-art methods, highlighting its effectiveness on graph anomaly detection tasks.

\vspace{0.5cm}
\textbf{Date}: May 26, 2026

\textbf{Author emails}: \email{\{tairanhuang\}@csu.edu.cn}, \email{\{qiangchen.sh\}@gmail.com}

\textbf{Correspondence}: \email{\{xiusu1994\}@csu.edu.cn}, \email{\{yichen\}@ust.hk}

\textbf{\faGithub~~Code}: \url{https://github.com/Tairan-Terrian/SignGAD}
\end{abstract}

\maketitle

\begin{figure}[h]
  \centering
  \includegraphics[width=1\linewidth]{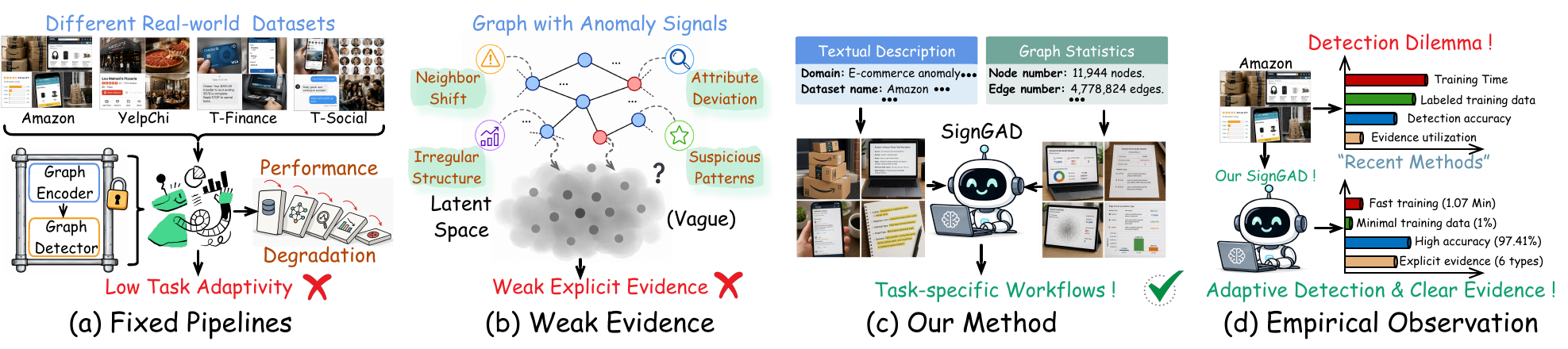}
  \caption{(a-b) Conceptual illustrations of the fundamental limitations in current GAD paradigms: \textit{fixed pipelines} and \textit{weak evidence}.
  (c) briefly describes our proposed method.
  As evidenced in (d), our method maintains impressive efficiency and delivers robust performance.
  }
  \label{motivation}
\end{figure}

\section{Introduction}
Graph anomaly detection (GAD) has been widely applied to many real-world detection scenarios, such as social networks \cite{GAD_1,social}, e-commerce platforms \cite{ecommerce_1,ecommerce}, and financial systems \cite{financial,financial_1}. 
These scenarios are naturally represented as attributed graphs, where relational structures encode interactions among entities and node attributes provide task-specific behavioral signals \cite{attad,miti,GAD_2}. 
The dominant paradigm in GAD is to exploit graph context for detecting anomaly nodes, since anomalies are often reflected not only by isolated attribute irregularities but also by their inconsistency with surrounding neighborhoods \cite{Fraud_1,GNN_1}. 
However, real-world graph anomalies are usually rare and expensive to annotate, which makes supervised detection highly constrained by limited labeled data \cite{Uniform,GAD,Sun}. 
Therefore, it is an important step to develop reliable GAD methods under limited supervision toward robust graph-based detection systems.


However, existing graph anomaly detection methods still face two key challenges, as shown in Figure \ref{motivation}. 
First, \textit{fixed pipelines} make current methods insufficiently task-adaptive, since the graph processing strategy and detector design are usually predefined before considering the specific task condition. 
This rigidity can lead to severe performance degradation under limited supervision, where only a few labeled nodes are available to correct an unsuitable detection workflow.
Second, \textit{weak evidence} makes existing methods less capable of explicitly leveraging anomaly-indicative graph signals. 
Although graph contexts often contain informative cues for distinguishing anomaly nodes, these cues are commonly absorbed into latent representations rather than organized as explicit detection evidence.
Consequently, the detector may fail to access reliable task-specific evidence with scarcely labeled anomalies.

To address these challenges, we propose Self-designing agentic workflows for few-shot Graph Anomaly Detection (SignGAD), a novel paradigm that shifts from fixed detector training to task-conditioned workflow-level design by automatically constructing detection processes.
Specifically, the \textit{Task-Conditioned Workflow Construction} introduces LLM-based agents to transform textual task descriptions and graph statistics into a structured task descriptor, thereby grounding workflow design in both semantic and structural task information.
The \textit{Evidence Graph Encoding} constructs contextual anomaly evidence, exposing node-level deviations to the detector instead of leaving them implicit in latent embeddings.
The \textit{Workflow Detector Bank} instantiates suitable detector agents for candidate workflows, enabling different graph anomaly detection tasks to benefit from different inductive biases under limited supervision.
The \textit{Validation Workflow Search} calibrates candidate workflows on validation nodes to select a task-suitable detection workflow, while the \textit{Guarded Final Refit} calibrates refit acceptance to further refine the selected workflow.

Our contributions are summarized as follows:
\begin{itemize}
    \item \textbf{Paradigm innovation.} 
    To the best of our knowledge, we are the first to reformulate graph anomaly detection from training a fixed anomaly detector to designing the task-conditioned detection workflows, which enables the detection process to be customized to different graph anomaly detection tasks.

    \item \textbf{Novel framework.} 
    We propose SignGAD, a self-designing agentic workflow framework for few-shot graph anomaly detection, which selects appropriate graph encodings and detector designs by constructing task-specific workflows and refines selected workflows through validation calibration and guarded final refit.

    \item \textbf{Performance validation.}
    We conduct extensive experiments on real-world graph anomaly detection datasets under limited supervision to verify the effectiveness of SignGAD, achieving state-of-the-art performance on graph anomaly detection tasks.
\end{itemize}

\begin{figure*}[th]
	\centering
	\includegraphics[width=\linewidth]{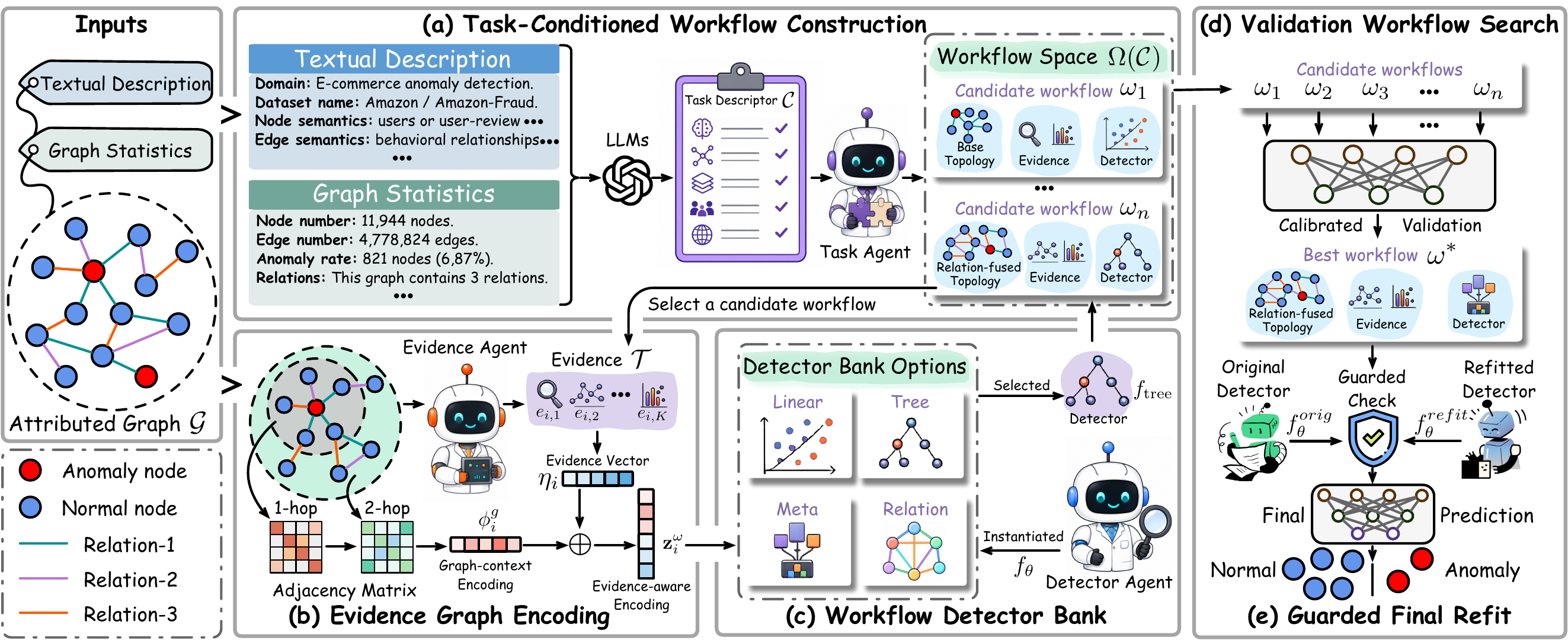}
	\caption{The overall architecture of SignGAD, including (a) Task-Conditioned Workflow Construction, (b) Evidence Graph Encoding, (c) Workflow Detector Bank, (d) Validation Workflow Search, and (e) Guarded Final Refit.
    }
	\label{framework}
\end{figure*}

\section{Related Work}
\subsection{Graph Anomaly Detection}
Existing graph anomaly detection methods mainly exploit graph structure and node attributes to identify anomaly nodes \cite{SLADE,GIF}. 
Traditional approaches rely on handcrafted statistics or shallow anomaly scores \cite{Cross,Deltagrad}, while recent Graph neural networks (GNNs) \cite{GCN} methods learn contextual node representations through neighborhood aggregation \cite{MVHGNN}. 
Other studies improve GAD by modifying graph structures \cite{homophily,SFGA}, using reconstruction objectives \cite{recon}, or considering heterogeneous \cite{HeteGAD} and dynamic graph settings \cite{SAD,SLADE}. 
More recently, large language models have been introduced to enhance GAD with semantic understanding and task-level reasoning \cite{MLED}. 
Despite these advances, most methods still follow predefined detection pipelines with fixed graph processing and detector designs. 
In contrast, SignGAD formulates GAD as a task-conditioned workflow design problem, explicitly organizing contextual evidence and selecting suitable detectors for different graph anomaly detection tasks.

\subsection{Agentic Workflow Design}
Agentic workflow design has recently attracted increasing attention for automating complex decision processes with large language models \cite{workflow,Agent4EDU}. 
Compared with conventional automation methods that search over fixed model configurations, agentic systems use LLMs as high-level planners to interpret task instructions \cite{task,GeoFlow} and organize tool-using procedures \cite{tool,flow}. 
Recent studies have applied this idea to scientific discovery and machine learning automation \cite{auto}, demonstrating the potential of LLM agents for structured decision making \cite{auto_1,tool_1}. 
However, most existing agentic workflows focus on general-purpose reasoning or open-ended tool use, while the agentic paradigm to graph anomaly detection remains unexplored.
SignGAD addresses this gap by grounding agentic planning in graph-specific detection workflows, where task information guides the construction of the detection process.

\section{Methodology}
\subsection{Preliminaries}
\noindent\textbf{Few-Shot Graph Anomaly Detection.}
We define the node-level graph anomaly detection under limited supervision as a few-shot scenario. 
Let $\mathcal{G}=(\mathcal{V},\{\mathcal{E}^{(r)}\}_{r=0}^{R},\mathbf{X})$ be an attributed graph, where $\mathcal{V}=\{v_1,\ldots,v_N\}$ is the node set, $\mathbf{X}\in\mathbb{R}^{N\times d}$ denotes node attributes, and $\mathcal{E}^{(r)}$ denotes the edge set of the $r$-th relation. 
Each node $v_i$ has a binary label $y_i\in\{0,1\}$, where $y_i=1$ indicates the anomaly node.
The node set is split into training subsets $\mathcal{V}_{\mathrm{tr}}$, validation subsets $\mathcal{V}_{\mathrm{val}}$, and test subsets $\mathcal{V}_{\mathrm{te}}$ :
\begin{equation}
\mathcal{V}
=
\mathcal{V}_{\mathrm{tr}}
\cup
\mathcal{V}_{\mathrm{val}}
\cup
\mathcal{V}_{\mathrm{te}},
\
\mathcal{V}_{\mathrm{tr}}
\cap
\mathcal{V}_{\mathrm{val}}
=
\mathcal{V}_{\mathrm{tr}}
\cap
\mathcal{V}_{\mathrm{te}}
=
\mathcal{V}_{\mathrm{val}}
\cap
\mathcal{V}_{\mathrm{te}}
=
\emptyset .
\end{equation}

On the few-shot graph anomaly detection scenario, only a small subset of labels in $\mathcal{V}_{\mathrm{tr}}$ is available for model training. 
The task is node-level binary classification, where each test node $v_i$ is assigned a predicted label:
\begin{equation}
\hat{y}_i\in\{0,1\},
\quad
v_i\in\mathcal{V}_{\mathrm{te}}.
\end{equation}
Here, $\hat{y}_i=1$ denotes an anomaly prediction and $\hat{y}_i=0$ denotes a normal prediction.

\noindent\textbf{Detection Workflow Design.}
Rather than assuming a fixed detector, we formulate graph anomaly detection as a workflow design problem. 
The detection workflow is defined as:
\begin{equation}
\omega
=
(g,\mathcal{T},f_{\theta},\tau),
\end{equation}
where $g$ denotes the selected graph topology, $\mathcal{T}$ denotes a set of anomaly evidence functions, $f_{\theta}$ denotes a trainable detector, and $\tau$ denotes a decision rule.
Given a workflow $\omega$, the detector induces a node-level binary classifier:
\begin{equation}
h_{\omega}
:
(\mathcal{G},v_i)
\mapsto
\hat{y}_i^{\omega},
\quad
\hat{y}_i^{\omega}\in\{0,1\}.
\end{equation}
The workflow therefore specifies not only how node representations and anomaly evidence are constructed, but also how the final binary decision is produced.

\noindent\textbf{Workflow Selection Objective.}
Let $\Omega(\mathcal{C})$ be the task-conditioned workflow space induced by task information $\mathcal{C}$, including textual task description and graph statistics. 
The objective is to select the workflow whose induced classifier performs best on validation nodes:
\begin{equation}
\omega^{*}
=
\arg\max_{\omega\in\Omega(\mathcal{C})}
\mathcal{Q}_{\mathrm{val}}
\left(
h_{\omega}
\right),
\end{equation}
where $\mathcal{Q}_{\mathrm{val}}(\cdot)$ denotes a validation criterion for binary graph anomaly detection.

\subsection{Task-Conditioned Workflow Construction}
The proposed SignGAD first employs task agents to construct a task-conditioned workflow space $\Omega(\mathcal{C})$ before fitting any detector.
Given an input graph $\mathcal{G}$, a small labeled set $\mathcal{V}_{\mathrm{tr}}$, and a textual task description $\mathcal{D}_{\mathrm{text}}$, SignGAD summarizes the task into a structured information descriptor:
\begin{equation}
\mathcal{C}
=
\Pi_{\mathrm{LLM}}
\left(
\mathcal{D}_{\mathrm{text}},
\mathrm{Stat}(\mathcal{G}),
\mathcal{V}_{\mathrm{tr}}
\right),
\end{equation}
where $\Pi_{\mathrm{LLM}}$ denotes LLM-based semantic planning and $\mathrm{Stat}(\mathcal{G})$ contains several graph statistics. 
The descriptor $\mathcal{C}$ serves as the interface between task understanding and workflow construction.
The task agent convert $\mathcal{C}$ into candidate workflow decisions, including the graph topology $g$, the evidence functions $\mathcal{T}$, and the detector family $f_{\theta}$. 
This process can be written as:
\begin{equation}
\Omega(\mathcal{C})
=
\mathrm{Plan}
\left(
\Omega_0;
\mathcal{C}
\right),
\end{equation}
where $\Omega_0$ denotes the initial workflow pool and $\mathrm{Plan}(\cdot)$ denotes the LLM-guided planning process. 
The planner instantiates task-compatible workflows according to the graph structure, relation information, supervision budget, and domain description.
For a single-relation graph, the graph component $g$ of a workflow corresponds to the original adjacency matrix $\mathbf{A}^{(0)}$. 
For a multi-relation graph, SignGAD additionally constructs a relation-fused adjacency matrix:
\begin{equation}
\mathbf{A}^{\mathrm{fused}}
=
\mathbb{I}
\left(
\mathbf{A}^{(0)}
+
\sum_{r=1}^{R}
\mathbf{A}^{(r)}
>
0
\right),
\end{equation}
where $\mathbf{A}^{(0)}$ is the base adjacency matrix and $\mathbf{A}^{(r)}$ denotes the adjacency matrix of the $r$-th additional relation. 
Thus, $g$ can instantiate either the base topology or the relation-fused topology, allowing the workflow space to adapt to both single-relation and multi-relation graphs.
This step changes graph anomaly detection from fitting a predefined detector into constructing a task-specific workflow space:
\begin{equation}
(\mathcal{G},\mathcal{V}_{\mathrm{tr}})
\rightarrow
f_{\theta}
\quad
\Longrightarrow
\quad
(\mathcal{G},\mathcal{V}_{\mathrm{tr}},\mathcal{D}_{\mathrm{text}},\mathcal{C})
\rightarrow
\Omega(\mathcal{C}).
\end{equation}
The constructed space is then used by subsequent modules for evidence encoding, detector instantiation, and validation-based workflow selection.

\subsection{Evidence Graph Encoding}
Given a workflow $\omega \in \Omega(\mathcal{C})$, this module introduces the evidence agent to encode each node under the selected graph topology $g$ and organize task-relevant anomaly signals.
Let $\mathbf{A}_{g}$ be the adjacency matrix induced by $g$, which can be the base topology $\mathbf{A}^{(0)}$ or the relation-fused topology $\mathbf{A}^{\mathrm{fused}}$.
We first compute the row-normalized adjacency matrix:
\begin{equation}
\widetilde{\mathbf{A}}_{g}
=
\mathbf{D}_{g}^{-1}
\mathbf{A}_{g},
\quad
(\mathbf{D}_{g})_{ii}
=
\sum_j
(\mathbf{A}_{g})_{ij}.
\end{equation}
Then SignGAD aggregates the first-order and the second-order neighborhood information:
\begin{equation}
\mathbf{H}^{(1)}_{g}
=
\widetilde{\mathbf{A}}_{g}\mathbf{X},
\quad
\mathbf{H}^{(2)}_{g}
=
\widetilde{\mathbf{A}}_{g}\mathbf{H}^{(1)}_{g}.
\end{equation}
For node $v_i$, its graph-context encoding is:
\begin{equation}
\begin{aligned}
\phi_i^{g}
=
\big[
&\mathbf{x}_i
\Vert
\log(1+d_i^{g})
\Vert
\mathbf{h}^{(1)}_{i,g}
\Vert
\mathbf{h}^{(2)}_{i,g}
\\
&\Vert
\left|
\mathbf{x}_i-\mathbf{h}^{(1)}_{i,g}
\right|
\Vert
\left|
\mathbf{h}^{(1)}_{i,g}-\mathbf{h}^{(2)}_{i,g}
\right|
\big],
\end{aligned}
\end{equation}
where $d_i^{g}$ is the degree of $v_i$ under topology $g$, and $\mathbf{h}^{(1)}_{i,g}$ and $\mathbf{h}^{(2)}_{i,g}$ are the $i$-th row vectors of $\mathbf{H}^{(1)}_{g}$ and $\mathbf{H}^{(2)}_{g}$, respectively.
The difference terms explicitly measure how much a node deviates from its local context.
The selected evidence functions provide task-relevant anomaly signals, whose detailed definitions are given in Appendix \ref{app:evidence}. 
For $\mathcal{T}=\{t_1,\ldots,t_K\}$, each function produces the scalar evidence score:
\begin{equation}
e_{i,k}
=
t_k
\left(
\mathcal{G},
g,
v_i
\right),
\quad
k=1,\ldots,K.
\end{equation}
The evidence vector of node $v_i$ is defined as:
\begin{equation}
\eta_i
=
\left[
e_{i,1}
\Vert
e_{i,2}
\Vert
\cdots
\Vert
e_{i,K}
\right].
\end{equation}
Finally, SignGAD forms the evidence-aware graph encoding:
\begin{equation}
\mathbf{z}_i^{\omega}
=
\left[
\phi_i^{g}
\Vert
\eta_i
\right].
\end{equation}
When $\mathcal{T}=\emptyset$, the evidence vector is omitted and $\mathbf{z}_i^{\omega}=\phi_i^{g}$. 
This encoding exposes both contextual deviations and explicit anomaly evidence to the subsequent detector.

\subsection{Workflow Detector Bank}
Given the evidence-aware encoding $\mathbf{z}_i^{\omega}$, the detector component $f_{\theta}$ is instantiated from a workflow detector bank through detector agents. 
Each detector agent corresponds to a candidate detector design and provides a specific inductive bias under limited supervision. 
Formally, the instantiated detector maps node encodings to intermediate anomaly scores:
\begin{equation}
s_i^{\omega}
=
f_{\theta}
\left(
\mathbf{z}_i^{\omega}
\right),
\quad
s_i^{\omega}\in[0,1].
\end{equation}
The first type of detector is a linear graph detector. 
It applies a class-balanced logistic classifier to the encoded node representation:
\begin{equation}
p_i
=
\sigma
\left(
\mathbf{w}^{\top}
\mathbf{z}_i^{\omega}
+
b
\right),
\end{equation}
where $\sigma(\cdot)$ is the sigmoid function. 
The output probability is used as the anomaly score $s_i^{\omega} = p_i$.
This detector is simple and stable when the labeled training set is small.
The second type of detector is a non-linear graph detector. 
It fits a tree-based classifier over the same encoded representation:
\begin{equation}
s_i^{\omega}
=
f_{\mathrm{tree}}
\left(
\mathbf{z}_i^{\omega}
\right),
\end{equation}
where $f_{\mathrm{tree}}$ denotes an ensemble tree model trained on $\mathcal{V}_{\mathrm{tr}}$. 
Compared with the linear detector, this model captures non-linear interactions between graph-context features and anomaly evidence.
To combine complementary detector behaviors, SignGAD further constructs stacked detector features. 
Let $\{f_1,\ldots,f_M\}$ be a set of base detectors selected from the bank. 
Their scores are concatenated as:
\begin{equation}
\psi_i^{\omega}
=
\left[
f_1(\mathbf{z}_i^{\omega})
\Vert
f_2(\mathbf{z}_i^{\omega})
\Vert
\cdots
\Vert
f_M(\mathbf{z}_i^{\omega})
\right].
\end{equation}
The stacked representation is then defined by $\rho_i^{\omega} = \left[ \mathbf{z}_i^{\omega} \Vert \psi_i^{\omega} \right]$.
A meta-detector is trained on $\rho_i^{\omega}$ to produce the final workflow score:
\begin{equation}
s_i^{\omega}
=
f_{\mathrm{meta}}
\left(
\rho_i^{\omega}
\right).
\end{equation}
This design allows the workflow to reuse decisions of multiple detectors as higher-level anomaly evidence.
For multi-relation graphs, the detector bank also supports relation-aware encoding. 
For each additional relation $r$, we construct a relation-specific representation $\phi_i^{(r)}$ using the same graph encoding procedure on $\mathbf{A}^{(r)}$. 
To keep the representation compact, we project it into a low-dimensional vector:
\begin{equation}
\mathbf{q}_i^{(r)}
=
\mathrm{Proj}_{r}
\left(
\phi_i^{(r)}
\right).
\end{equation}
The relation-specific detector feature is:
\begin{equation}
\chi_i^{(r)}
=
\left[
\mathbf{q}_i^{(r)}
\Vert
f_{\mathrm{lin}}^{(r)}
\left(
\phi_i^{(r)}
\right)
\Vert
f_{\mathrm{tree}}^{(r)}
\left(
\phi_i^{(r)}
\right)
\right],
\end{equation}
where $f_{\mathrm{lin}}^{(r)}$ and $f_{\mathrm{tree}}^{(r)}$ are relation-specific base detectors.
The relation-aware representation is obtained by concatenating relation-level features:
\begin{equation}
\xi_i^{\omega}
=
\left[
\rho_i^{\omega}
\Vert
\chi_i^{(1)}
\Vert
\cdots
\Vert
\chi_i^{(R)}
\right].
\end{equation}
The relation-aware detector outputs:
\begin{equation}
s_i^{\omega}
=
f_{\mathrm{rel}}
\left(
\xi_i^{\omega}
\right).
\end{equation}
When the input graph is single-relation, this relation-aware branch is skipped and the detector operates on $\rho_i^{\omega}$.

Overall, the detector bank enables each workflow to instantiate a detector that matches the task condition encoded in $\mathcal{C}$. 
The selected detector is trained only on $\mathcal{V}_{\mathrm{tr}}$ and produces the intermediate score used by the workflow classifier.

\subsection{Validation Workflow Search}
After evidence encoding and detector instantiation, each candidate workflow $\omega\in\Omega(\mathcal{C})$ produces an intermediate anomaly score $s_i^{\omega}$ for node $v_i$. 
We calibrate the decision threshold of each workflow on the validation set.
Here, $\Theta$ denotes the candidate threshold set.
For a candidate threshold $\tau\in\Theta$, workflow $\omega$ produces validation predictions:
\begin{equation}
\hat{y}_{i}^{\omega}(\tau)
=
\mathbb{I}
\left(
s_i^{\omega}
\ge
\tau
\right),
\quad
v_i\in\mathcal{V}_{\mathrm{val}}.
\end{equation}
The threshold of workflow $\omega$ is selected by maximizing validation F1-Macro:
\begin{equation}
\tau_{\omega}^{*}
=
\arg\max_{\tau\in\Theta}
\mathrm{F1}_{\mathrm{macro}}
\left(
y_{\mathrm{val}},
\hat{y}_{\mathrm{val}}^{\omega}(\tau)
\right).
\end{equation}
With the selected threshold, the calibrated validation predictions are $\hat{y}_{\mathrm{val}}^{\omega}
=
\hat{y}_{\mathrm{val}}^{\omega}
\left(
\tau_{\omega}^{*}
\right)$.
Let $A_{\omega}$ denote the validation AUC computed from the continuous scores, and let $F_{\omega}$ denote the validation F1-Macro computed from binary predictions:
\begin{equation}
A_{\omega}
=
\mathrm{AUC}
\left(
y_{\mathrm{val}},
s_{\mathrm{val}}^{\omega}
\right),
\quad
F_{\omega}
=
\mathrm{F1}_{\mathrm{macro}}
\left(
y_{\mathrm{val}},
\hat{y}_{\mathrm{val}}^{\omega}
\right).
\end{equation}
To avoid selecting unnecessarily complex workflows under limited supervision, we further introduce a complexity penalty $\Omega_{\mathrm{c}}(\omega)$. 
The validation criterion is defined as:
\begin{equation}
\mathcal{Q}_{\mathrm{val}}(\omega)
=
\left[
\min(A_{\omega},F_{\omega}),
\;
\alpha A_{\omega}
+
(1-\alpha)F_{\omega},
\;
-\Omega_{\mathrm{c}}(\omega)
\right],
\end{equation}
where $\alpha\in[0,1]$ controls the balance between ranking quality and binary classification quality. 
The final workflow is selected as:
\begin{equation}
\omega^{*}
=
\arg\max_{\omega\in\Omega(\mathcal{C})}
\mathcal{Q}_{\mathrm{val}}(\omega).
\end{equation}

\subsection{Guarded Final Refit}
After the workflow $\omega^{*}$ is selected, SignGAD optionally refits only its detector component with additional labeled nodes. 
This step does not change the evidence functions or decision rule. 
It only uses part of the validation labels to enhance detector fitting under limited supervision.
To avoid unsafe refitting, we split the validation set into refit subset and calibration subset:
\begin{equation}
\mathcal{V}_{\mathrm{val}}
=
\mathcal{V}_{\mathrm{refit}}
\cup
\mathcal{V}_{\mathrm{cal}},
\quad
\mathcal{V}_{\mathrm{refit}}
\cap
\mathcal{V}_{\mathrm{cal}}
=
\emptyset .
\end{equation}
The refit subset is combined with the original training set:
\begin{equation}
\mathcal{V}_{\mathrm{tr}}^{+}
=
\mathcal{V}_{\mathrm{tr}}
\cup
\mathcal{V}_{\mathrm{refit}}.
\end{equation}
Let $f_{\theta}^{\mathrm{orig}}$ be the detector trained on $\mathcal{V}_{\mathrm{tr}}$, and let $f_{\theta}^{\mathrm{refit}}$ be the detector retrained on $\mathcal{V}_{\mathrm{tr}}^{+}$. 
Both detectors are evaluated on $\mathcal{V}_{\mathrm{cal}}$ using the same selected workflow $\omega^{*}$. 
We denote their calibration performance by:
\begin{equation}
m_{\mathrm{orig}}
=
\left(
A_{\mathrm{orig}},
F_{\mathrm{orig}}
\right),
\quad
m_{\mathrm{refit}}
=
\left(
A_{\mathrm{refit}},
F_{\mathrm{refit}}
\right),
\end{equation}
where $A$ and $F$ denote AUC and F1-Macro on $\mathcal{V}_{\mathrm{cal}}$, respectively.
The refitted detector is accepted only when it is not worse than the original detector on the calibration subset. 
We compare the rounded metric pairs lexicographically:
\begin{equation}
\mathrm{round}
\left(
m_{\mathrm{refit}},
4
\right)
\ge_{\mathrm{lex}}
\mathrm{round}
\left(
m_{\mathrm{orig}},
4
\right).
\end{equation}
If the condition holds, $f_{\theta}^{\mathrm{refit}}$ replaces $f_{\theta}^{\mathrm{orig}}$ in the selected workflow.
Otherwise, SignGAD keeps the original detector. 
After workflow selection and guarded refit, the calibrated classifier induced by $\omega^{*}$ is used for final prediction:
\begin{equation}
\hat{y}_i
=
h_{\omega^{*}}
\left(
\mathcal{G},
v_i
\right),
\quad
v_i\in\mathcal{V}_{\mathrm{te}}.
\end{equation}

\section{Experiments}
\subsection{Experimental Setup}
\noindent{\bfseries Datasets.} We use the Amazon \cite{amazon}, YelpChi \cite{yelp}, T-Finance \cite{BWGNN}, and T-Social \cite{BWGNN} datasets to evaluate the performance of SignGAD. 
All the datasets consist of users and their associated comments.
The detailed statistics of the datasets are provided in the Appendix \ref{app:dataset}.

\noindent{\bfseries Baselines.} 
We conduct a comprehensive comparison of SignGAD with eighteen baselines, which fall into two categories: the generic GNN models and the graph anomaly detection models.
The generic GNN models include MLP \cite{MLP}, GCN \cite{GCN}, GraphSAGE \cite{GraphSAGE}, GAT \cite{GAT}, GIN \cite{GIN}, and GATv2 \cite{GATv2}.
The graph anomaly detection models include BWGNN \cite{BWGNN}, GHRN \cite{GHRN}, GDN \cite{GDN}, GAGA \cite{GAGA}, DiG-In-GNN \cite{DiG}, ConsisGAD \cite{consisGAD},  SpaceGNN \cite{SpaceGNN}, and TAQ-GAD \cite{TAQ-GAD}.
All the baselines are implemented using the source codes released by the authors.
The details of these baseline models are introduced in the Appendix \ref{app:baseline}.

\begin{table*}[]
\centering
\caption{Overall performance (\%) on four datasets with the 1\% training ratio on two evaluation metrics, where the reported metrics are AUC and F1-Macro. Highlighted are the results ranked \best{first}, \second{second}, and \third{third}.}
\resizebox{1\linewidth}{!}{
\begin{tabular}{ccccccccc}
\toprule
Datasets                          & \multicolumn{2}{c}{Amazon}                         & \multicolumn{2}{c}{YelpChi}                       & \multicolumn{2}{c}{T-Finance}                      & \multicolumn{2}{c}{T-Social}                    \\ \cmidrule(lr){2-3} \cmidrule(lr){4-5} \cmidrule(lr){6-7} \cmidrule(lr){8-9} 
Metric                          & AUC           & F1-Macro                           & AUC          & F1-Macro                           & AUC           & F1-Macro                           & AUC          & F1-Macro                         \\ \midrule
 MLP          & 92.39 ± 0.72  &  87.53 ± 1.61  & 72.18 ± 0.39 &  61.61 ± 0.33  & 92.17 ± 0.64  &  82.33 ± 0.54  & 66.95 ± 0.71 & 54.09 ± 0.61                     \\
 GCN          & 87.34 ± 0.59  &  70.94 ± 2.43  & 54.65 ± 0.53 &  35.59 ± 10.27 & 89.29 ± 0.19  &  77.16 ± 1.20  & 83.30 ± 1.60 & 65.16 ± 0.92                     \\
 GraphSAGE    & 90.12 ± 0.48  &  84.25 ± 2.26  & 73.70 ± 0.52 &  63.33 ± 0.51  & 89.42 ± 1.36  &  77.62 ± 1.87  & 71.45 ± 2.24 & \multicolumn{1}{l}{56.47 ± 0.64} \\
 GAT          & 80.74 ± 3.64  &  63.45 ± 12.82 & 70.14 ± 1.91 &  61.22 ± 1.32  & 87.40 ± 4.41  &  75.49 ± 5.63  & 73.46 ± 3.32 & 61.98 ± 2.06                     \\
 GIN          & 84.35 ± 0.75  &  71.20 ± 1.37  & 56.98 ± 0.82 &  53.58 ± 0.41  & 81.29 ± 1.66  &  65.38 ± 3.05  & 78.70 ± 2.19 & 61.62 ± 5.93                     \\
 GATv2        & 85.39 ± 3.19  &  76.20 ± 13.69 & 72.83 ± 0.75 &  62.23 ± 0.56  & 73.25 ± 10.00 &  63.16 ± 9.02  & 79.89 ± 4.80 & 62.99 ± 1.34                     \\ \midrule
 CARE-GNN     & 89.68 ± 0.76  &  75.74 ± 0.50  & 72.11 ± 1.23 &  61.62 ± 0.87  & 91.45 ± 0.40  &  83.68 ± 0.78  & - OOM -       & - OOM -                          \\
 GraphConsis  & 64.23 ± 13.83 &  55.35 ± 8.09  & 80.13 ± 0.88 &  64.96 ± 2.54  & 92.61 ± 0.47  &  85.37 ± 0.48  & - OOM -       & - OOM -                          \\
 PC-GNN       & 91.18 ± 0.66  &  85.25 ± 2.09  & 75.17 ± 0.44 &  64.23 ± 0.47  & 91.74 ± 0.85  &  86.97 ± 0.24  & 64.68 ± 0.64 & 49.66 ± 0.12                     \\
 BWGNN        & 88.56 ± 0.87  &  \third{90.48 ± 0.98}  & 77.62 ± 2.37 &  66.54 ± 0.73  & 93.08 ± 1.57  &  86.96 ± 1.51  & 84.40 ± 3.01 & 76.37 ± 1.82                     \\
 H2-Fdetector & 83.81 ± 2.57  &  72.46 ± 4.01  & 72.38 ± 1.13 &  63.97 ± 0.66  & - OOM -       &  - OOM -       & - OOM -       & - OOM -                          \\
 GHRN         & 88.35 ± 2.03  &  86.35 ± 2.60  & 75.33 ± 1.44 &  63.62 ± 1.41  & 91.93 ± 0.93  &  80.05 ± 4.43  & 84.20 ± 3.91 & 71.25 ± 4.32                     \\
 GDN          & 92.43 ± 0.12  &  89.75 ± 0.05  & 75.92 ± 0.51 &  64.81 ± 0.25  & 88.75 ± 1.79  &  76.62 ± 3.90  & 67.69 ± 1.49 & 55.76 ± 0.82                     \\
 GAGA         & 82.61 ± 6.87  &  76.85 ± 8.08  & 71.61 ± 2.13 &  61.81 ± 1.69  & 92.36 ± 1.45  &  81.10 ± 2.60  & 78.92 ± 1.26 & 65.58 ± 3.30                     \\
 DiG-In-GNN   & 86.30 ± 0.67  &  82.61 ± 0.21  & 74.40 ± 0.93 &  63.36 ± 0.27  & 92.53 ± 0.31  &  88.51 ± 0.12  & 93.75 ± 0.45 & \third{81.40 ± 0.78}             \\ 
 ConsisGAD    & 93.91 ± 0.58  &  90.03 ± 0.53  & \third{83.36 ± 0.53} &  69.72 ± 0.30  & 95.33 ± 0.30  &  \second{90.97 ± 0.63}  & 94.31 ± 0.20 & 78.08 ± 0.54                     \\
 SpaceGNN     & \third{94.02 ± 0.27}  &  \second{90.69 ± 0.66}  & 82.47 ± 0.67 &  \third{70.21 ± 0.56}  & \third{95.84 ± 0.57}  &  \best{91.21 ± 0.73}  & \third{95.20 ± 1.03} & 77.34 ± 0.62                     \\
 TAQ-GAD      & \second{95.71 ± 0.40}  &  90.22 ± 0.34  & \second{86.95 ± 0.75} &  \second{75.49 ± 0.41}  & \second{96.02 ± 0.54}  &  90.11 ± 0.42  & \second{96.58 ± 0.85} & \second{85.46 ± 0.69}            \\
\midrule
 SignGAD      & \best{97.41 ± 0.54}  &  \best{91.17 ± 0.35}  & \best{92.68 ± 0.84}  &  \best{81.55 ± 0.37}  & \best{96.79 ± 0.48}  &  \third{90.42 ± 0.26}  & \best{98.37 ± 0.25}  & \best{93.72 ± 0.61}              \\ \bottomrule
\end{tabular}
}
\label{tab:results}
\end{table*}

\begin{table*}[t]
\centering

\begin{minipage}[t]{0.48\textwidth}
\centering
\caption{F1-Macro comparison with different LLM variants.}
\resizebox{\linewidth}{!}{
\begin{tabular}{lcccc}
\toprule
Variants & Amazon & YelpChi & T-Finance & T-Social \\
\midrule
\textit{LLaMA3-8B} & 89.41 ± 0.45 & 79.75 ± 0.65 & 88.65 ± 0.52 & 91.95 ± 0.72 \\
\textit{GPT-3}     & \third{89.98 ± 0.55} & \third{80.32 ± 0.51} & \third{89.31 ± 0.44} & \third{92.58 ± 0.68} \\
\textit{GPT-3.5}   & \second{90.62 ± 0.41} & \second{81.05 ± 0.42} & \second{89.95 ± 0.31} & \second{93.25 ± 0.55} \\
\textit{GPT-4.1}   & \best{91.17 ± 0.35} & \best{81.55 ± 0.37} & \best{90.42 ± 0.26} & \best{93.72 ± 0.61} \\
\bottomrule
\end{tabular}
}
\label{tab:llm_variants}
\end{minipage}
\hfill
\begin{minipage}[t]{0.48\textwidth}
\centering
\caption{Comparison on Amazon dataset with F1-Macro under different training ratios.}
\resizebox{\linewidth}{!}{
\begin{tabular}{lcccc}
\toprule
Methods & 1\% & 3\% & 5\% & 10\% \\
\midrule
ConsisGAD 
& 90.03 ± 0.53 
& 91.26 ± 0.42 
& 91.88 ± 0.33 
& 92.11 ± 0.26 \\
SpaceGNN  
& \second{90.69 ± 0.66} 
& \third{91.57 ± 0.29} 
& \second{92.41 ± 0.43} 
& \second{92.76 ± 0.37} \\
TAQ-GAD   
& \third{90.22 ± 0.34} 
& \second{91.83 ± 0.51}
& \third{92.14 ± 0.23} 
& \third{92.59 ± 0.17} \\
SignGAD   
& \best{91.17 ± 0.35} 
& \best{92.54 ± 0.27} 
& \best{93.19 ± 0.21} 
& \best{93.51 ± 0.14} \\
\bottomrule
\end{tabular}
}
\label{tab:train_ratio_amazon}
\end{minipage}

\end{table*}

\begin{table*}[t]
\centering

\begin{minipage}[t]{0.48\textwidth}
\centering
\caption{Ablation study of SignGAD under the 1\% training ratio. The reported metric is F1-Macro.}
\resizebox{\linewidth}{!}{
\begin{tabular}{lcccc}
\toprule
Variants & Amazon & YelpChi & T-Finance & T-Social \\
\midrule
\textit{w/o LLM}
& 89.89 ± 0.42 & 80.08 ± 0.45 & 89.04 ± 0.33 & 92.41 ± 0.69 \\
\textit{w/o Evid.}
& 89.51 ± 0.47 & 79.62 ± 0.42 & 89.21 ± 0.31 & 91.87 ± 0.73 \\
\textit{w/o Bank}
& 89.74 ± 0.39 & 80.31 ± 0.48 & 88.58 ± 0.36 & 92.23 ± 0.65 \\
\textit{w/o Search}
& 89.42 ± 0.51 & 79.84 ± 0.53 & 88.75 ± 0.39 & 92.05 ± 0.76 \\
\textit{w/o Refit}
& 88.43 ± 0.55 & 78.82 ± 0.59 & 87.83 ± 0.45 & 90.84 ± 0.85 \\
\midrule
SignGAD
& \best{91.17 ± 0.35} & \best{81.55 ± 0.37} & \best{90.42 ± 0.26} & \best{93.72 ± 0.61} \\
\bottomrule
\end{tabular}
}
\label{tab:ablation}
\end{minipage}
\hfill
\begin{minipage}[t]{0.48\textwidth}
\centering
\caption{Evidence function number analysis with F1-Macro.}
\resizebox{\linewidth}{!}{
\begin{tabular}{ccccc}
\toprule
$K$ & Amazon & YelpChi & T-Finance & T-Social \\
\midrule
0 & 89.15 ± 0.49 & 78.52 ± 0.53 & 88.24 ± 0.41 & 91.35 ± 0.84 \\
1 & 89.84 ± 0.42 & 79.46 ± 0.48 & 88.93 ± 0.37 & 92.08 ± 0.77 \\
2 & 90.41 ± 0.39 & 80.31 ± 0.44 & 89.56 ± 0.33 & 92.81 ± 0.71 \\
3 & \second{90.92 ± 0.37} & \third{81.08 ± 0.39} & \third{90.11 ± 0.29} & \third{93.35 ± 0.65} \\
4 & \best{91.17 ± 0.35} & \best{81.55 ± 0.37} & \best{90.42 ± 0.26} & \best{93.72 ± 0.61} \\
5 & \third{90.87 ± 0.41} & \second{81.12 ± 0.42} & \second{90.15 ± 0.31} & \second{93.41 ± 0.67} \\
6 & 90.31 ± 0.47 & 80.47 ± 0.49 & 89.48 ± 0.38 & 92.86 ± 0.75 \\
\bottomrule
\end{tabular}
}
\label{tab:sens_evidence_num}
\end{minipage}

\end{table*}

\begin{table*}[t]
\centering

\begin{minipage}[t]{0.48\textwidth}
\centering
\caption{Analysis of candidate workflow number $N_{\omega}$. The reported metric is F1-Macro.}
\resizebox{1\linewidth}{!}{
\begin{tabular}{ccccc}
\toprule
$N_{\omega}$ & Amazon & YelpChi & T-Finance & T-Social \\
\midrule
5  
& 89.24 ± 0.48 
& 78.43 ± 0.62 
& 87.89 ± 0.51 
& 91.27 ± 0.86 \\
10 
& \third{90.31 ± 0.41} 
& \third{80.12 ± 0.48} 
& \third{89.34 ± 0.38} 
& \third{92.58 ± 0.72} \\
20 
& \best{91.17 ± 0.35} 
& \best{81.55 ± 0.37} 
& \second{90.42 ± 0.26} 
& \best{93.72 ± 0.61} \\
30 
& \second{91.15 ± 0.37} 
& \second{81.52 ± 0.39} 
& \best{90.45 ± 0.28} 
& \second{93.68 ± 0.63} \\
\bottomrule
\end{tabular}
}
\label{tab:sens_workflow_num}
\end{minipage}
\hfill
\begin{minipage}[t]{0.48\textwidth}
\centering
\caption{Analysis of validation balance coefficient $\alpha$. The reported metric is F1-Macro.}
\resizebox{1\linewidth}{!}{
\begin{tabular}{ccccc}
\toprule
$\alpha$ & Amazon & YelpChi & T-Finance & T-Social \\
\midrule
0.00 
& 88.52 ± 0.51 
& 78.41 ± 0.56 
& 87.89 ± 0.42 
& 91.31 ± 0.78 \\
0.25 
& \third{90.15 ± 0.43} 
& \third{80.33 ± 0.45} 
& \third{89.47 ± 0.34} 
& \third{92.84 ± 0.68} \\
0.50 
& \best{91.17 ± 0.35} 
& \best{81.55 ± 0.37} 
& \best{90.42 ± 0.26} 
& \best{93.72 ± 0.61} \\
0.75 
& \second{90.64 ± 0.42} 
& \second{80.92 ± 0.42} 
& \second{89.86 ± 0.31} 
& \second{93.25 ± 0.64} \\
1.00 
& 89.23 ± 0.56 
& 79.18 ± 0.53 
& 88.54 ± 0.39 
& 92.14 ± 0.73 \\
\bottomrule
\end{tabular}
}
\label{tab:sens_alpha}
\end{minipage}

\end{table*}

\noindent{\bfseries Experimental Settings.}
For all baselines, we conduct a 10-run experiment to evaluate their performance.
On the few-shot graph anomaly detection scenario, we set the training ratio to 1\% and use GPT-4.1 as the default LLM backbone through an API service for workflow planning.
The overall performance is evaluated with two metrics including AUC \cite{Yili}  and F1-Macro \cite{F1}. 



\begin{figure*}[th]
	\centering
    \includegraphics[width=\linewidth]{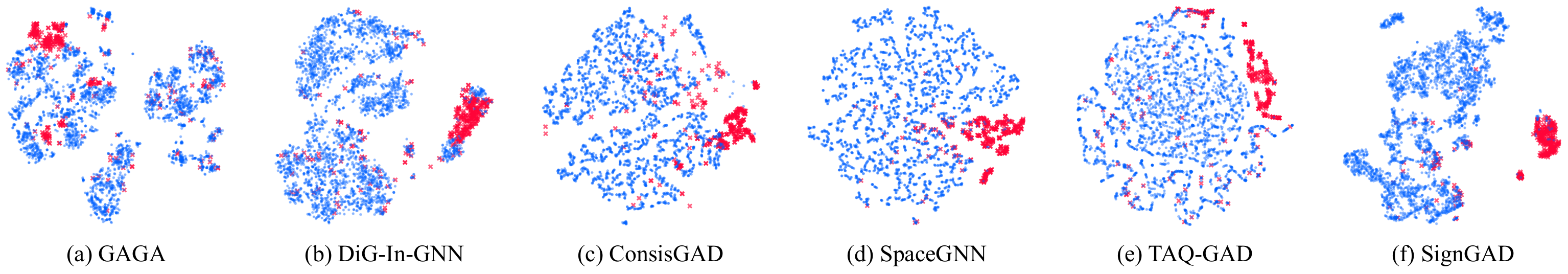}
	\caption{The visualization analysis on the Amazon dataset, where red represents fraudsters blue represents benign entities.}
	\label{visualization}
\end{figure*}

\begin{table}[t]
\centering
\caption{Evidence selection frequency across datasets.}
\resizebox{0.7\linewidth}{!}{
\begin{tabular}{lcccc}
\toprule
Evidence & Amazon & YelpChi & T-Finance & T-Social \\
\midrule
Degree Anomaly & 0.75 & 0.68 & 0.54 & 0.71 \\
Neighbor Feature Deviation & 0.91 & 0.88 & 0.76 & 0.83 \\
Feature Smoothness & 0.82 & 0.63 & 0.48 & 0.69 \\
Feature Reconstruction Residual & 0.79 & 0.84 & 0.72 & 0.66 \\
Relation Degree Profile & 0.57 & 0.57 & -- & -- \\
Relation Disagreement & 0.61 & 0.61 & -- & -- \\
\bottomrule
\end{tabular}
}
\label{tab:evidence_frequency}
\end{table}

\begin{figure}[th]
	\centering
    \includegraphics[width=0.7\linewidth]{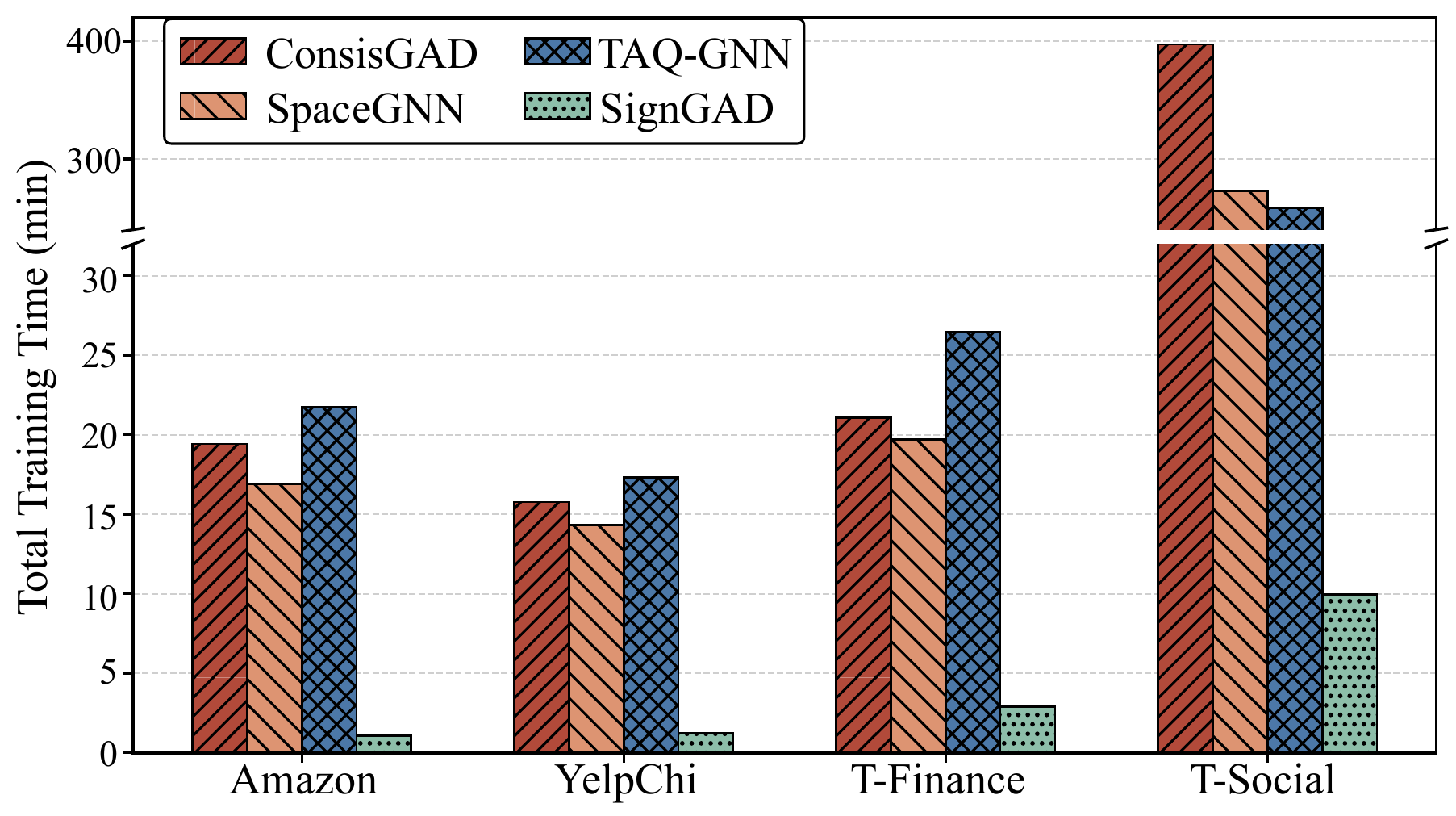}
	\caption{The time cost comparison on four datasets.}
	\label{training_time}
\end{figure}

\subsection{Experimental Results}
\noindent{\bfseries Main Results.} 
Table \ref{tab:results} reports the performance of the state-of-the-art methods of our proposed SignGAD on four real-world datasets. 
SignGAD achieves the best performance on seven out of eight evaluation cases, demonstrating its strong effectiveness under limited supervision. 
These results indicate that the proposed task-conditioned workflow design can better adapt to different graph anomaly detection tasks than fixed detector architectures. 
Overall, SignGAD constructs more suitable detection processes for different graph tasks, verifying the necessity of workflow-level adaptation in few-shot graph anomaly detection.

\noindent{\bfseries LLM Variants.} 
Table~\ref{tab:llm_variants} reports the results of SignGAD with different LLM variants on F1-Macro, with the complete AUC and F1-Macro results provided in Appendix \ref{app:llm_var}. 
GPT-4.1 consistently achieves the best performance across all datasets, while GPT-3.5 ranks second, indicating that stronger planning ability leads to more suitable workflow construction. 
Nevertheless, all variants maintain competitive performance, showing that SignGAD is not tied to a specific LLM backbone. 
These results verify the importance of agentic planning in constructing effective detection workflows for few-shot graph anomaly detection.

\noindent{\bfseries Varying Supervision.} 
Table \ref{tab:train_ratio_amazon} reports the F1-Macro results on Amazon dataset under different training ratios, with the complete results on all datasets provided in Appendix \ref{app:vary}. 
SignGAD consistently achieves the best performance across all ratios, showing that the proposed workflow design remains effective under different supervision levels. 
The advantage is maintained even at the 1\% training ratio, indicating that task-conditioned agentic workflows can better exploit contextual anomaly evidence when labeled nodes are limited.

\noindent{\bfseries Ablation Study.} 
Table \ref{tab:ablation} reports the ablation results of SignGAD under the 1\% training ratio, with the full results and detailed description provided in Appendix \ref{app:ablation}. 
Removing any component consistently degrades performance, verifying that each module contributes to the final workflow. 
Among all variants, \textit{w/o Refit} causes the largest drop, showing the importance of guarded final refit for refining the selected workflow. 
The drops of \textit{w/o Evid.} and \textit{w/o Search} further demonstrate that explicit anomaly evidence and validation-based workflow calibration are both necessary for reliable few-shot graph anomaly detection.

\subsection{Additional Analysis}
\noindent{\bfseries Analysis of Evidence Function Number.} 
Table \ref{tab:sens_evidence_num} reports the sensitivity analysis of the evidence function number $K$, with the full results provided in Appendix \ref{app:function}. 
The performance consistently improves when increasing $K$ from 0 to 4, showing that explicit anomaly evidence helps the detector capture task-relevant graph signals. 
The best results are achieved at $K=4$ across all datasets, while further increasing $K$ slightly degrades performance. 
This indicates that a compact evidence set is sufficient for effective detection, whereas excessive evidence may introduce redundant or noisy signals under limited supervision.

\noindent{\bfseries Analysis of Candidate Workflow Number.} 
Table \ref{tab:sens_workflow_num} reports the sensitivity of SignGAD to the candidate workflow number $N_{\omega}$, with the full results provided in Appendix \ref{app:candidate}. 
Increasing $N_{\omega}$ from 5 to 20 consistently improves performance, indicating that a richer workflow pool helps validation search identify more suitable detection processes. 
When $N_{\omega}$ further increases to 30, the performance becomes saturated and remains close to that of $N_{\omega}=20$. 
Therefore, we set $N_{\omega}=20$ as the default setting to balance detection performance and workflow search cost.

\noindent{\bfseries Analysis of Validation Balance Coefficient.} 
Table \ref{tab:sens_alpha} reports the sensitivity of SignGAD to the validation balance coefficient $\alpha$, with the full results provided in Appendix \ref{app:validation}. 
The best performance is consistently achieved at $\alpha=0.50$, showing that balancing ranking quality and binary prediction quality is important for workflow selection. 
When $\alpha$ is set to either 0 or 1, the performance drops noticeably, indicating that relying on a single validation objective is insufficient for robust few-shot graph anomaly detection. 
Therefore, we set $\alpha=0.50$ as the default value.

\noindent{\bfseries Efficiency Analysis.} 
Figure \ref{training_time} reports the time cost comparison on four datasets. 
SignGAD achieves substantially lower runtime than all compared methods across all datasets, reducing the cost from tens or even hundreds of minutes to only a few minutes. 
The advantage is especially clear on T-Social, where SignGAD only takes 9.98 minutes, while the strongest baselines require more than 250 minutes. 
This efficiency mainly comes from the workflow-level design of SignGAD, which constructs task-conditioned detection processes with lightweight evidence encoding and detector instantiation instead of repeatedly training heavy graph neural detectors. 
These results show that SignGAD is not only effective under limited supervision, but also practical for large-scale graph anomaly detection.

\noindent{\bfseries Visualization Analysis.} 
To illustrate the effectiveness of the proposed method in a more intuitive way, we conduct the visualization task on the Amazon dataset.
We compare several baselines with the visualization results of the proposed method SignGAD, as shown in Figure \ref{visualization}.
The results indicate that SignGAD enables anomaly nodes to form compact clusters that are more clearly separated from normal nodes.
The visualization experiments demonstrate that SignGAD enhances the discriminative capability of graph anomaly detection through agentic workflows.
The clear separation between anomaly and normal nodes further validates the effectiveness of SignGAD in realistic scenarios.

\noindent{\bfseries Evidence Contribution Analysis.} 
Table \ref{tab:evidence_frequency} reports the selection frequency of different evidence functions across datasets. 
Neighbor Feature Deviation is selected most frequently on all datasets, showing that neighborhood-level attribute inconsistency provides a generally useful signal for identifying anomaly nodes. 
Meanwhile, different datasets rely on different evidence patterns, such as the higher frequency of Feature Smoothness on Amazon and the higher frequency of Feature Reconstruction Residual on YelpChi. 
These results indicate that SignGAD can adjust evidence usage according to task characteristics, supporting the necessity of task-conditioned evidence construction in graph anomaly detection.

\section{Conclusion}
In this paper, we proposed SignGAD, a self-designing agentic workflow framework for few-shot graph anomaly detection. 
SignGAD shifts graph anomaly detection from fixed detector training to task-conditioned workflow-level design, enabling the detection process to be customized to different graph anomaly detection tasks. 
Through workflow construction, SignGAD determines suitable graph encodings and detector designs to exploit task-specific anomaly evidence. 
The guarded final refit strategy further refines the selected workflow through refit acceptance calibration, improving reliability under limited supervision. 
Extensive experiments demonstrate that SignGAD achieves state-of-the-art performance on real-world datasets, and further analyses confirm the necessity of workflow-level customization.

\clearpage

\bibliography{ref}

\clearpage
\section*{Appendix}
\appendix
\renewcommand{\appendixname}{\appendixname~\Alph{section}}

\section{Evidence Function Definitions}
\label{app:evidence}
This section provides detailed definitions of the evidence functions used in SignGAD. 
Each evidence function produces a node-level scalar signal that reflects a specific type of anomaly-indicative pattern. 
Given a workflow $\omega\in\Omega(\mathcal{C})$, the selected evidence set is denoted as $\mathcal{T}=\{t_1,\ldots,t_K\}$. 
For node $v_i$, each evidence function outputs:
\begin{equation}
e_{i,k}
=
t_k
\left(
\mathcal{G},
g,
v_i
\right),
\quad
k=1,\ldots,K.
\end{equation}
The evidence scores are normalized into a comparable range before being used by the detector:
\begin{equation}
\bar{e}_{i,k}
=
\frac{
e_{i,k}
-
\min_{v_j\in\mathcal{V}} e_{j,k}
}{
\max_{v_j\in\mathcal{V}} e_{j,k}
-
\min_{v_j\in\mathcal{V}} e_{j,k}
+
\epsilon
},
\end{equation}
where $\epsilon$ is a small constant for numerical stability. 
The normalized evidence vector is:
\begin{equation}
\eta_i
=
\left[
\bar{e}_{i,1}
\Vert
\bar{e}_{i,2}
\Vert
\cdots
\Vert
\bar{e}_{i,K}
\right].
\end{equation}

\subsection{Degree Anomaly}
Degree anomaly measures whether a node has an unusual connectivity pattern compared with the labeled training nodes. 
Let $d_i^g$ denote the degree of node $v_i$ under the selected topology $g$. 
We compute the mean and standard deviation of node degrees on the training set:
\begin{equation}
\mu_d
=
\frac{1}{|\mathcal{V}_{\mathrm{tr}}|}
\sum_{v_j\in\mathcal{V}_{\mathrm{tr}}}
d_j^g,
\quad
\sigma_d
=
\sqrt{
\frac{1}{|\mathcal{V}_{\mathrm{tr}}|}
\sum_{v_j\in\mathcal{V}_{\mathrm{tr}}}
\left(
d_j^g-\mu_d
\right)^2
}.
\end{equation}
The degree anomaly evidence is defined as the absolute z-score:
\begin{equation}
t_{\mathrm{deg}}
\left(
\mathcal{G},
g,
v_i
\right)
=
\left|
\frac{
d_i^g-\mu_d
}{
\sigma_d+\epsilon
}
\right|.
\end{equation}
A larger value indicates that the node has an unusually high or low degree compared with the training distribution.

\subsection{Relation Degree Profile}
For multi-relation graphs, anomalous nodes may show inconsistent activity across relations. 
Let $\mathbf{A}^{(0)}$ be the base adjacency matrix and $\mathbf{A}^{(r)}$ be the adjacency matrix of the $r$-th additional relation. 
The degree of node $v_i$ under relation $r$ is:
\begin{equation}
d_i^{(r)}
=
\sum_j
\mathbf{A}_{ij}^{(r)}.
\end{equation}
We construct the relation degree profile:
\begin{equation}
\mathbf{p}_i
=
\left[
\bar{d}_i^{(0)}
\Vert
\bar{d}_i^{(1)}
\Vert
\cdots
\Vert
\bar{d}_i^{(R)}
\right],
\end{equation}
where $\bar{d}_i^{(r)}$ denotes the normalized degree of node $v_i$ under relation $r$.
The relation degree profile evidence is:
\begin{equation}
t_{\mathrm{rdp}}
\left(
\mathcal{G},
g,
v_i
\right)
=
\mathrm{Std}
\left(
\mathbf{p}_i
\right).
\end{equation}
A larger value indicates that the node exhibits highly uneven connectivity across different relations. 
For single-relation graphs, this evidence function is omitted.

\subsection{Relation Disagreement}
Relation disagreement measures the inconsistency between the base topology and the additional relation structures. 
It is designed for multi-relation graphs where abnormal nodes may behave normally in one relation but suspiciously in another.
We first compute the normalized degree under the base relation and the average normalized degree under additional relations:
\begin{equation}
\bar{d}_i^{\mathrm{add}}
=
\frac{1}{R}
\sum_{r=1}^{R}
\bar{d}_i^{(r)}.
\end{equation}
The relation disagreement evidence is defined as:
\begin{equation}
t_{\mathrm{rel}}
\left(
\mathcal{G},
g,
v_i
\right)
=
\left|
\bar{d}_i^{(0)}
-
\bar{d}_i^{\mathrm{add}}
\right|.
\end{equation}
A larger value indicates that the node has inconsistent connectivity between the base relation and the additional relations. 
For single-relation graphs, this evidence function is omitted.

\subsection{Neighbor Feature Deviation}
Neighbor feature deviation measures how much a node attribute differs from the average attribute of its graph neighborhood. 
Let $\widetilde{\mathbf{A}}_g=\mathbf{D}_g^{-1}\mathbf{A}_g$ be the row-normalized adjacency matrix under topology $g$. 
The first-order neighborhood representation is:
\begin{equation}
\mathbf{h}_{i,g}^{(1)}
=
\sum_j
\widetilde{\mathbf{A}}_{g,ij}
\mathbf{x}_j.
\end{equation}
The neighbor feature deviation evidence is:
\begin{equation}
t_{\mathrm{nfd}}
\left(
\mathcal{G},
g,
v_i
\right)
=
\left\|
\mathbf{x}_i
-
\mathbf{h}_{i,g}^{(1)}
\right\|_2.
\end{equation}
A larger value indicates that the node is less consistent with its local attribute context.

\subsection{Feature Smoothness}
Feature smoothness measures local feature inconsistency using the average absolute deviation between a node and its neighborhood. 
Using the same neighborhood representation $\mathbf{h}_{i,g}^{(1)}$, the smoothness evidence is:
\begin{equation}
t_{\mathrm{smooth}}
\left(
\mathcal{G},
g,
v_i
\right)
=
\frac{1}{d}
\left\|
\mathbf{x}_i
-
\mathbf{h}_{i,g}^{(1)}
\right\|_1,
\end{equation}
where $d$ is the feature dimension.
Compared with neighbor feature deviation, this evidence emphasizes dimension-wise deviations and is less sensitive to a few extremely large feature differences.

\subsection{Feature Reconstruction Residual}
Feature reconstruction residual captures attribute-space abnormality by measuring how well a node attribute can be reconstructed from a low-dimensional feature subspace. 
We fit a truncated singular value decomposition on the training node attributes:
\begin{equation}
\mathbf{X}_{\mathrm{tr}}
\approx
\mathbf{U}
\mathbf{\Sigma}
\mathbf{V}^{\top}.
\end{equation}
Let $\mathrm{Proj}(\cdot)$ and $\mathrm{Rec}(\cdot)$ denote the projection and reconstruction functions induced by the fitted low-rank subspace. 
The reconstructed attribute of node $v_i$ is:
\begin{equation}
\widehat{\mathbf{x}}_i
=
\mathrm{Rec}
\left(
\mathrm{Proj}
\left(
\mathbf{x}_i
\right)
\right).
\end{equation}

The reconstruction residual evidence is:
\begin{equation}
t_{\mathrm{rec}}
\left(
\mathcal{G},
g,
v_i
\right)
=
\left\|
\mathbf{x}_i
-
\widehat{\mathbf{x}}_i
\right\|_2.
\end{equation}
A larger value indicates that the node is poorly explained by the dominant attribute patterns in the training set.

\subsection{Evidence Usage}
The evidence functions above capture complementary anomaly patterns from structural connectivity, neighborhood context, relation inconsistency, and attribute reconstruction. 
Given the selected evidence set $\mathcal{T}$, SignGAD concatenates their normalized outputs into $\eta_i$ and combines it with the graph-context encoding:
\begin{equation}
\mathbf{z}_i^{\omega}
=
\left[
\phi_i^g
\Vert
\eta_i
\right].
\end{equation}
When no evidence function is selected, SignGAD uses only the graph-context encoding:
\begin{equation}
\mathbf{z}_i^{\omega}
=
\phi_i^g.
\end{equation}

\begin{table}[]
\centering
\caption{Statistics of the datasets.}
\resizebox{0.5\linewidth}{!}{
\begin{tabular}{ccccc}
    \toprule
    \multicolumn{1}{l}{\textbf{Dataset}} & \multicolumn{1}{l}{\textbf{\#Nodes}} & \multicolumn{1}{l}{\textbf{Anomaly(\%)}}  & \textbf{Relation} & \textbf{\#Edges} \\ \midrule
    \multirow{3}{*}{Amazon}  & \multirow{3}{*}{11,944}  & \multirow{3}{*}
    {\begin{tabular}[c]{@{}c@{}}821\\ (6.87\%)\end{tabular}}  
    & U-P-U  & 175,608 \\
    & & & U-S-U & 3,566,479 \\
    & & & U-V-U & 1,036,737 \\
    \midrule
    \multirow{3}{*}{YelpChi} & \multirow{3}{*}{45,954}  & \multirow{3}{*}
    {\begin{tabular}[c]{@{}c@{}}6,677\\ (14.53\%)\end{tabular}}
    & R-U-R & 49,315  \\
    & & & R-T-R  & 573,616           \\
    & & & R-S-R & 3,402,743        \\
        \midrule
        T-Finance & 39,357 & {\begin{tabular}[c]{@{}c@{}}1,804\\ (4.58\%)\end{tabular}} & Transaction  & 21,222,543 \\
        \midrule
        T-Social & 5,781,065 & {\begin{tabular}[c]{@{}c@{}}174,280\\ (3.01\%)\end{tabular}} & Interaction   & 73,105,508 \\
    \bottomrule
\end{tabular}
}
\label{tab:datasets}
\end{table}

\begin{figure*}[th]
	\centering
    \includegraphics[width=\linewidth]{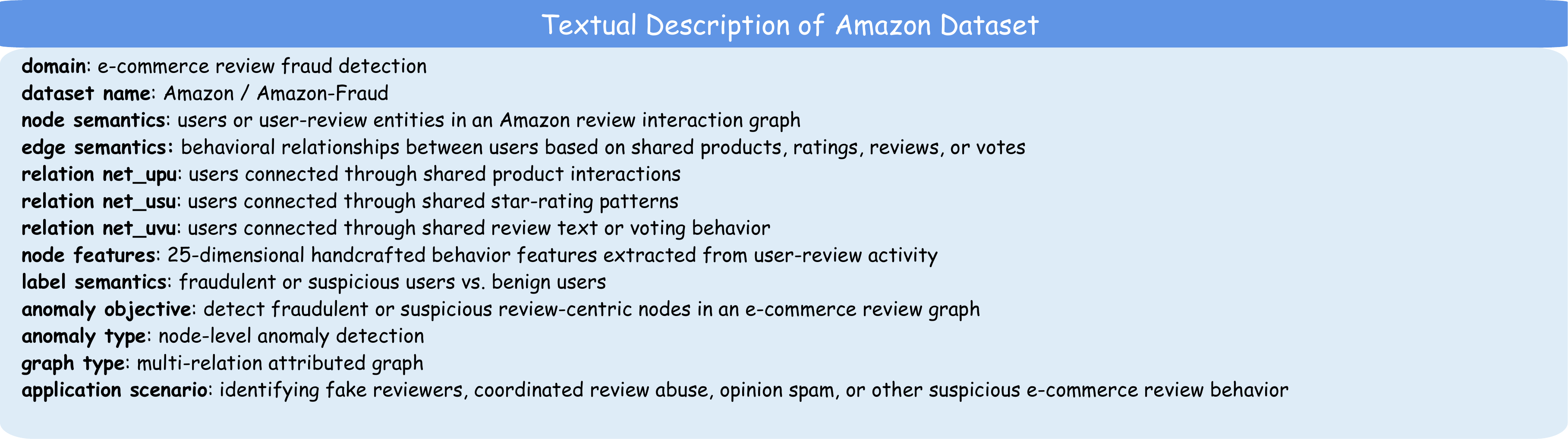}
	\caption{The detailed textual description of Amazon dataset.}
	\label{amazon}
\end{figure*}

\begin{figure*}[th]
	\centering
    \includegraphics[width=\linewidth]{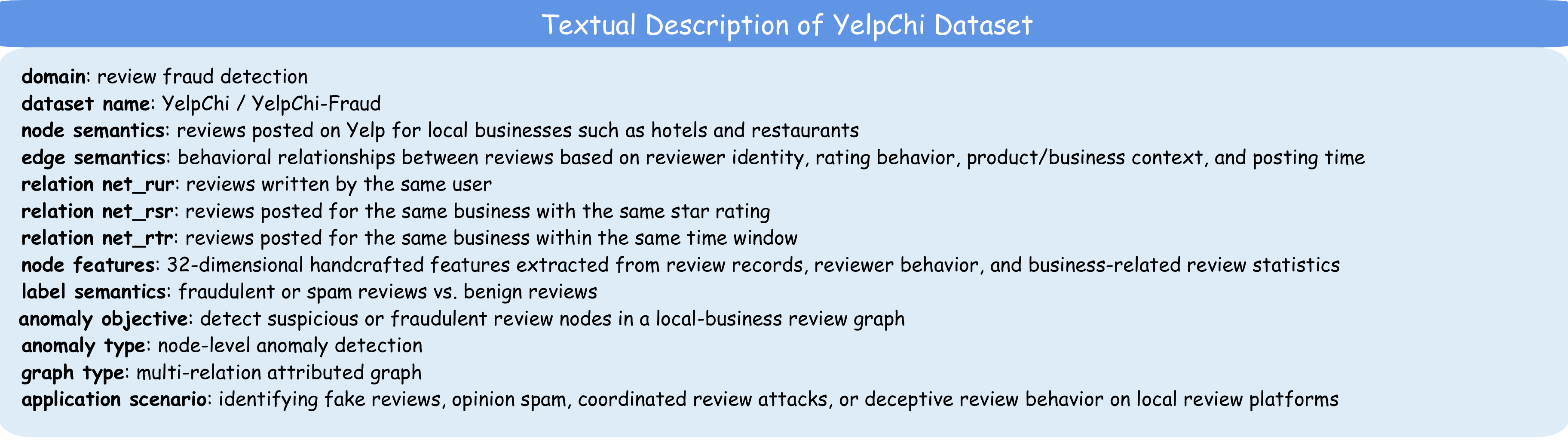}
	\caption{The detailed textual description of YelpChi dataset.}
	\label{yelp}
\end{figure*}

\begin{figure*}[th]
	\centering
    \includegraphics[width=\linewidth]{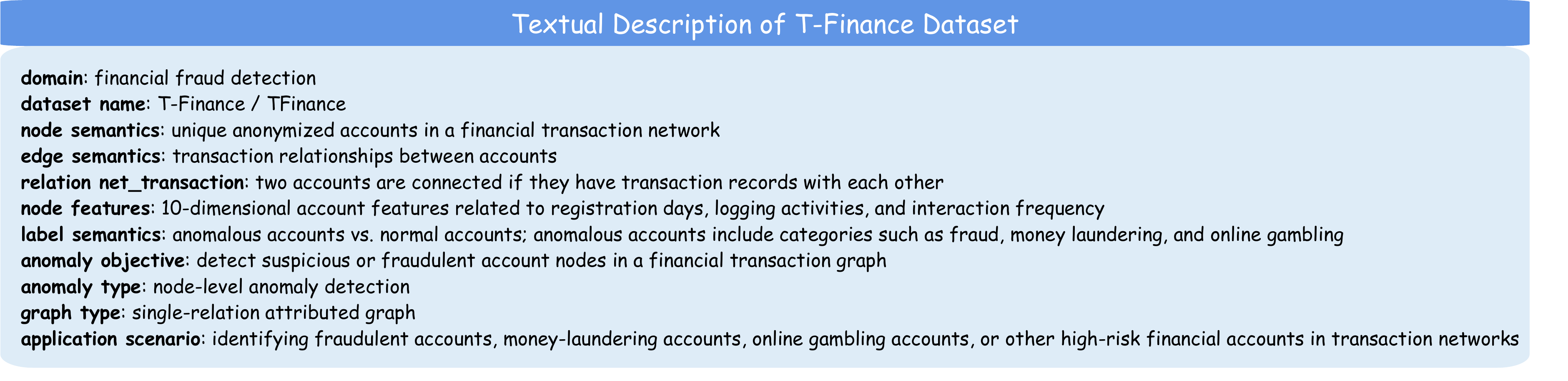}
	\caption{The detailed textual description of T-Finance dataset.}
	\label{finance}
\end{figure*}

\begin{figure*}[th]
	\centering
    \includegraphics[width=\linewidth]{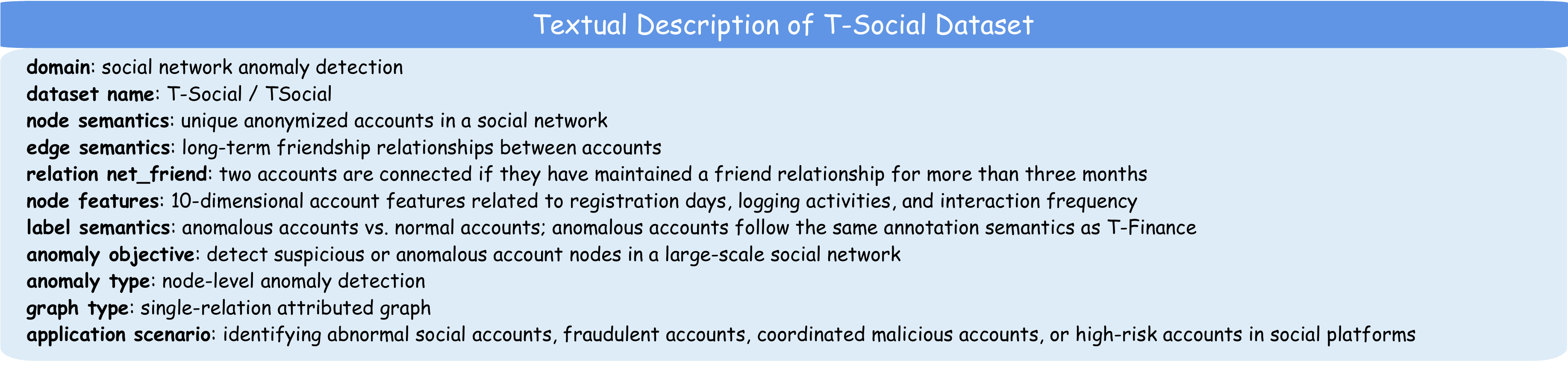}
	\caption{The detailed textual description of T-Social dataset.}
	\label{social}
\end{figure*}
\section{Datasets}
\label{app:dataset}
The four benchmark graph anomaly detection datasets are used for our empirical study: Amazon \cite{amazon}, YelpChi \cite{yelp}, T-Finance \cite{BWGNN}, and T-Social \cite{BWGNN}. 
Table \ref{tab:datasets} summarizes the statistics of these datasets. 
We offer comprehensive descriptions of each dataset as follows:
\begin{itemize}
	\item \textbf{Amazon \cite{amazon}} aims to identify fraudsters who post fake comments on products in categories of musical instruments.
    The detailed textual description is shown in Figure \ref{amazon}.
    The dataset contains three distinct types of relations between nodes in Amazon: 
    (1) U-P-U represents the connections between users who have reviewed one of the same products at least.
    (2) U-S-U refers to the connections between users who have given one identical star rating within the same week at least.
    (3) U-V-U denotes the connections between users whose mutual comment text similarity is in the top 5\%.
	\item \textbf{YelpChi \cite{yelp}} aims to identify fraud comments for hotels and restaurants in Chicago.
    The detailed textual description is shown in Figure \ref{yelp}.
    The YelpChi dataset includes three types of relations between nodes:
    (1) R-U-R refers to the connection between comments released by the same user.
    (2) R-T-R represents to the connection between comments that have same star rating for same product.
    (3) R-S-R denotes the connection between comments of same product released in same month.
	\item \textbf{T-Finance \cite{BWGNN}} is intended to detect the fraud accounts in transaction networks.
    The detailed textual description is shown in Figure \ref{finance}.
    The connections in the graph indicate pairs of accounts with transaction records between them.
    \item \textbf{T-Social \cite{BWGNN}} aims to identify anomalous users in social networks. 
    The detailed textual description is shown in Figure \ref{social}.
    The two users are connected if they have been friends for more than three months. 
\end{itemize}

\begin{table*}[t]
\centering
\caption{Performance comparison of SignGAD with different LLM variants under the 1\% training ratio. The reported metrics are AUC and F1-Macro. Highlighted are the results ranked \best{first}, \second{second}, and \third{third}.}
\resizebox{1\linewidth}{!}{
\begin{tabular}{lcccccccc}
\toprule
\multirow{2}{*}{Variants} 
& \multicolumn{2}{c}{Amazon} 
& \multicolumn{2}{c}{YelpChi} 
& \multicolumn{2}{c}{T-Finance} 
& \multicolumn{2}{c}{T-Social} \\
\cmidrule(lr){2-3} \cmidrule(lr){4-5} \cmidrule(lr){6-7} \cmidrule(lr){8-9}
& AUC & F1-Macro 
& AUC & F1-Macro 
& AUC & F1-Macro 
& AUC & F1-Macro \\
\midrule
LLaMA3-8B   & 95.68 ± 0.71 & 89.41 ± 0.45 & 90.82 ± 0.88 & 79.75 ± 0.65 & 95.05 ± 0.55 & 88.65 ± 0.52 & 96.65 ± 0.38 & 91.95 ± 0.72 \\
GPT-3       & \third{96.25 ± 0.62} & \third{89.98 ± 0.55} & \third{91.45 ± 0.91} & \third{80.32 ± 0.51} & \third{95.65 ± 0.61} & \third{89.31 ± 0.44} & \third{97.15 ± 0.45} & \third{92.58 ± 0.68} \\
GPT-3.5     & \second{96.95 ± 0.48} & \second{90.62 ± 0.41} & \second{92.12 ± 0.75} & \second{81.05 ± 0.42} & \second{96.31 ± 0.52} & \second{89.95 ± 0.31} & \second{97.92 ± 0.32} & \second{93.25 ± 0.55} \\
GPT-4.1     & \best{97.41 ± 0.54} & \best{91.17 ± 0.35} 
            & \best{92.68 ± 0.84} & \best{81.55 ± 0.37} 
            & \best{96.79 ± 0.48} & \best{90.42 ± 0.26} 
            & \best{98.37 ± 0.25} & \best{93.72 ± 0.61} \\
\bottomrule
\end{tabular}
}
\label{app:llm_variants}
\end{table*}

\begin{table*}[t]
\centering
\caption{Performance comparison under different training ratios. The reported metrics are AUC and F1-Macro with mean and standard deviation over 10 runs. Highlighted are the results ranked \best{first}, \second{second}, and \third{third}.}
\resizebox{1\linewidth}{!}{
\begin{tabular}{llcccccccc}
\toprule
\multirow{2}{*}{Datasets} 
& \multirow{2}{*}{Methods}
& \multicolumn{2}{c}{1\%}
& \multicolumn{2}{c}{3\%}
& \multicolumn{2}{c}{5\%}
& \multicolumn{2}{c}{10\%} \\
\cmidrule(lr){3-4}
\cmidrule(lr){5-6}
\cmidrule(lr){7-8}
\cmidrule(lr){9-10}
& & AUC & F1-Macro & AUC & F1-Macro & AUC & F1-Macro & AUC & F1-Macro \\
\midrule

\multirow{4}{*}{Amazon}
& ConsisGAD 
& 93.91 ± 0.58 & 90.03 ± 0.53 
& 95.14 ± 0.47 & 91.26 ± 0.42 & 95.82 ± 0.39 & 91.88 ± 0.33 & 96.17 ± 0.28 & 92.11 ± 0.26 \\
& SpaceGNN  
& \third{94.02 ± 0.27} & \second{90.69 ± 0.66} 
& \third{95.29 ± 0.24} &  \third{91.57 ± 0.29} & \third{96.08 ± 0.19} & \second{92.41 ± 0.43} & \third{96.44 ± 0.16} & \second{92.76 ± 0.37} \\
& TAQ-GAD   
& \second{95.71 ± 0.40} & \third{90.22 ± 0.34} 
& \second{96.86 ± 0.31} & \second{91.83 ± 0.51} & \second{97.43 ± 0.26} & \third{92.14 ± 0.23} & \second{97.82 ± 0.18} & \third{92.59 ± 0.17} \\
& SignGAD   
& \best{97.41 ± 0.54} & \best{91.17 ± 0.35} 
& \best{98.23 ± 0.43} & \best{92.54 ± 0.27} & \best{98.66 ± 0.34} & \best{93.19 ± 0.21} & \best{98.89 ± 0.22} & \best{93.51 ± 0.14} \\
\midrule

\multirow{4}{*}{YelpChi}
& ConsisGAD 
& \third{83.36 ± 0.53} & 69.72 ± 0.30 
& \third{86.51 ± 0.46} & 73.27 ± 0.27 & \third{88.13 ± 0.37} & 75.14 ± 0.21 & \third{89.26 ± 0.29} & 76.32 ± 0.16 \\
& SpaceGNN  
& 82.47 ± 0.67 & \third{70.21 ± 0.56} 
& 85.83 ± 0.53 & \third{74.08 ± 0.48} & 87.56 ± 0.41 & \third{76.21 ± 0.39} & 88.64 ± 0.33 & \third{77.57 ± 0.28} \\
& TAQ-GAD   
& \second{86.95 ± 0.75} & \second{75.49 ± 0.41} 
& \second{89.57 ± 0.61} & \second{78.63 ± 0.34} & \second{91.06 ± 0.47} & \second{80.24 ± 0.29} & \second{91.83 ± 0.38} & \second{81.16 ± 0.22} \\
& SignGAD   
& \best{92.68 ± 0.84} & \best{81.55 ± 0.37} 
& \best{94.13 ± 0.68} & \best{84.31 ± 0.31} & \best{95.27 ± 0.52} & \best{85.86 ± 0.24} & \best{95.84 ± 0.39} & \best{86.52 ± 0.19} \\
\midrule

\multirow{4}{*}{T-Finance}
& ConsisGAD 
& 95.33 ± 0.30 & \second{90.97 ± 0.63} 
& 96.51 ± 0.26 & \second{92.14 ± 0.51} & 97.16 ± 0.21 & \third{92.83 ± 0.38} & 97.47 ± 0.17 & \third{93.26 ± 0.29} \\
& SpaceGNN  
& \third{95.84 ± 0.57} & \best{91.21 ± 0.73} 
& \third{96.86 ± 0.43} & \best{92.47 ± 0.58} & \third{97.33 ± 0.34} & \second{93.11 ± 0.44} & \third{97.69 ± 0.27} & \second{93.54 ± 0.31} \\
& TAQ-GAD   
& \second{96.02 ± 0.54} & 90.11 ± 0.42 
& \second{97.13 ± 0.42} & 91.86 ± 0.35 & \second{97.64 ± 0.33} & 92.67 ± 0.26 & \second{97.96 ± 0.24} & 93.13 ± 0.19 \\
& SignGAD   
& \best{96.79 ± 0.48} & \third{90.42 ± 0.26} 
& \best{97.83 ± 0.37} & \third{92.27 ± 0.22} & \best{98.34 ± 0.29} & \best{93.29 ± 0.17} & \best{98.66 ± 0.21} & \best{93.83 ± 0.13} \\
\midrule

\multirow{4}{*}{T-Social}
& ConsisGAD 
& 94.31 ± 0.20 & \third{78.08 ± 0.54} 
& 96.07 ± 0.17 & \third{81.53 ± 0.44} & 96.86 ± 0.13 & \third{83.27 ± 0.36} & 97.23 ± 0.09 & \third{84.56 ± 0.27} \\
& SpaceGNN  
& \third{95.20 ± 1.03} & 77.34 ± 0.62 
& \third{96.63 ± 0.78} & 81.16 ± 0.53 & \third{97.37 ± 0.58} & 83.09 ± 0.41 & \third{97.74 ± 0.44} & 84.26 ± 0.32 \\
& TAQ-GAD   
& \second{96.58 ± 0.85} & \second{85.46 ± 0.69} 
& \second{97.87 ± 0.64} & \second{88.23 ± 0.54} & \second{98.44 ± 0.48} & \second{89.67 ± 0.43} & \second{98.76 ± 0.37} & \second{90.57 ± 0.31} \\
& SignGAD   
& \best{98.37 ± 0.25} & \best{93.72 ± 0.61} 
& \best{98.91 ± 0.21} & \best{95.14 ± 0.49} & \best{99.26 ± 0.16} & \best{95.83 ± 0.37} & \best{99.44 ± 0.11} & \best{96.36 ± 0.26} \\

\bottomrule
\end{tabular}
}
\label{app:train_ratio_full}
\end{table*}

\begin{table*}[t]
\centering
\caption{Full ablation study of SignGAD under the 1\% training ratio. The reported metrics are AUC and F1-Macro with mean and standard deviation over 10 runs.}
\resizebox{1\linewidth}{!}{
\begin{tabular}{lcccccccc}
\toprule
\multirow{2}{*}{Variants}
& \multicolumn{2}{c}{Amazon}
& \multicolumn{2}{c}{YelpChi}
& \multicolumn{2}{c}{T-Finance}
& \multicolumn{2}{c}{T-Social} \\
\cmidrule(lr){2-3}
\cmidrule(lr){4-5}
\cmidrule(lr){6-7}
\cmidrule(lr){8-9}
& AUC & F1-Macro
& AUC & F1-Macro
& AUC & F1-Macro
& AUC & F1-Macro \\
\midrule
\textit{w/o LLM}
& 96.08 ± 0.61 & 89.89 ± 0.42 
& 91.43 ± 0.91 & 80.08 ± 0.45 
& 95.31 ± 0.56 & 89.04 ± 0.33 
& 96.75 ± 0.32 & 92.41 ± 0.69 \\
\textit{w/o Evid.}
& 95.67 ± 0.65 & 89.51 ± 0.47 
& 90.87 ± 0.96 & 79.62 ± 0.42 
& 95.65 ± 0.52 & 89.21 ± 0.31 
& 96.93 ± 0.35 & 91.87 ± 0.73 \\
\textit{w/o Bank}
& 96.32 ± 0.58 & 89.74 ± 0.39 
& 91.21 ± 0.89 & 80.31 ± 0.48 
& 95.12 ± 0.58 & 88.58 ± 0.36 
& 97.14 ± 0.29 & 92.23 ± 0.65 \\
\textit{w/o Search}
& 95.53 ± 0.72 & 89.42 ± 0.51 
& 91.02 ± 0.94 & 79.84 ± 0.53 
& 95.43 ± 0.63 & 88.75 ± 0.39 
& 96.52 ± 0.41 & 92.05 ± 0.76 \\
\textit{w/o Refit}
& 94.76 ± 0.79 & 88.43 ± 0.55 
& 89.91 ± 1.05 & 78.82 ± 0.59 
& 94.27 ± 0.71 & 87.83 ± 0.45 
& 95.68 ± 0.48 & 90.84 ± 0.85 \\
\midrule
SignGAD
& \textbf{97.41 ± 0.54} & \textbf{91.17 ± 0.35}
& \textbf{92.68 ± 0.84} & \textbf{81.55 ± 0.37}
& \textbf{96.79 ± 0.48} & \textbf{90.42 ± 0.26}
& \textbf{98.37 ± 0.25} & \textbf{93.72 ± 0.61} \\
\bottomrule
\end{tabular}
}
\label{app:ablation_full}
\end{table*}

\begin{table*}[t]
\centering
\caption{Parameter sensitivity of evidence function number $K$. The reported metrics are AUC and F1-Macro.}
\resizebox{1\linewidth}{!}{
\begin{tabular}{ccccccccc}
\toprule
\multirow{2}{*}{$K$}
& \multicolumn{2}{c}{Amazon}
& \multicolumn{2}{c}{YelpChi}
& \multicolumn{2}{c}{T-Finance}
& \multicolumn{2}{c}{T-Social} \\
\cmidrule(lr){2-3}
\cmidrule(lr){4-5}
\cmidrule(lr){6-7}
\cmidrule(lr){8-9}
& AUC & F1-Macro
& AUC & F1-Macro
& AUC & F1-Macro
& AUC & F1-Macro \\
\midrule
0 & 95.37 ± 0.61 & 89.15 ± 0.49 & 89.61 ± 0.98 & 78.52 ± 0.53 & 94.75 ± 0.65 & 88.24 ± 0.41 & 96.12 ± 0.41 & 91.35 ± 0.84 \\
1 & 96.02 ± 0.58 & 89.84 ± 0.42 & 90.54 ± 0.93 & 79.46 ± 0.48 & 95.31 ± 0.59 & 88.93 ± 0.37 & 96.74 ± 0.38 & 92.08 ± 0.77 \\
2 & 96.67 ± 0.55 & 90.41 ± 0.39 & 91.38 ± 0.89 & 80.31 ± 0.44 & 95.94 ± 0.54 & 89.56 ± 0.33 & 97.41 ± 0.34 & 92.81 ± 0.71 \\
3 & \third{97.15 ± 0.51} & \second{90.92 ± 0.37} & \third{92.15 ± 0.86} & \third{81.08 ± 0.39} & \third{96.42 ± 0.51} & \third{90.11 ± 0.29} & \third{98.05 ± 0.29} & \third{93.35 ± 0.65} \\
4 & \best{97.41 ± 0.54} & \best{91.17 ± 0.35} & \best{92.68 ± 0.84} & \best{81.55 ± 0.37} & \best{96.79 ± 0.48} & \best{90.42 ± 0.26} & \best{98.37 ± 0.25} & \best{93.72 ± 0.61} \\
5 & \second{97.23 ± 0.59} & \third{90.87 ± 0.41} & \second{92.41 ± 0.91} & \second{81.12 ± 0.42} & \second{96.53 ± 0.55} & \second{90.15 ± 0.31} & \second{98.14 ± 0.31} & \second{93.41 ± 0.67} \\
6 & 96.85 ± 0.66 & 90.31 ± 0.47 & 91.82 ± 1.03 & 80.47 ± 0.49 & 95.98 ± 0.62 & 89.48 ± 0.38 & 97.68 ± 0.39 & 92.86 ± 0.75 \\
\bottomrule
\end{tabular}
}
\label{app:sens_evidence_num_full}
\end{table*}

\begin{table*}[t]
\centering
\caption{Parameter sensitivity of candidate workflow number $N_{\omega}$. The reported metrics are AUC and F1-Macro.}
\resizebox{1\linewidth}{!}{
\begin{tabular}{ccccccccc}
\toprule
\multirow{2}{*}{$N_{\omega}$}
& \multicolumn{2}{c}{Amazon}
& \multicolumn{2}{c}{YelpChi}
& \multicolumn{2}{c}{T-Finance}
& \multicolumn{2}{c}{T-Social} \\
\cmidrule(lr){2-3}
\cmidrule(lr){4-5}
\cmidrule(lr){6-7}
\cmidrule(lr){8-9}
& AUC & F1-Macro
& AUC & F1-Macro
& AUC & F1-Macro
& AUC & F1-Macro \\
\midrule
5  
& 95.82 ± 0.71 
& 89.24 ± 0.48 
& 89.91 ± 1.12 
& 78.43 ± 0.62 
& 94.63 ± 0.75 
& 87.89 ± 0.51 
& 96.48 ± 0.53 
& 91.27 ± 0.86 \\
10 
& \third{96.65 ± 0.62} 
& \third{90.31 ± 0.41} 
& \third{91.35 ± 0.95} 
& \third{80.12 ± 0.48} 
& \third{95.81 ± 0.59} 
& \third{89.34 ± 0.38} 
& \third{97.52 ± 0.38} 
& \third{92.58 ± 0.72} \\
20 
& \second{97.41 ± 0.54} 
& \best{91.17 ± 0.35} 
& \second{92.68 ± 0.84} 
& \best{81.55 ± 0.37} 
& \best{96.79 ± 0.48} 
& \second{90.42 ± 0.26} 
& \second{98.37 ± 0.25} 
& \best{93.72 ± 0.61} \\
30 
& \best{97.43 ± 0.52} 
& \second{91.15 ± 0.37} 
& \best{92.71 ± 0.82} 
& \second{81.52 ± 0.39} 
& \second{96.77 ± 0.49} 
& \best{90.45 ± 0.28} 
& \best{98.39 ± 0.26} 
& \second{93.68 ± 0.63} \\
\bottomrule
\end{tabular}
}
\label{app:sens_workflow_num_full}
\end{table*}

\begin{table*}[t]
\centering
\caption{Parameter sensitivity of validation balance coefficient $\alpha$. The reported metrics are AUC and F1-Macro.}
\resizebox{1\linewidth}{!}{
\begin{tabular}{ccccccccc}
\toprule
\multirow{2}{*}{$\alpha$}
& \multicolumn{2}{c}{Amazon}
& \multicolumn{2}{c}{YelpChi}
& \multicolumn{2}{c}{T-Finance}
& \multicolumn{2}{c}{T-Social} \\
\cmidrule(lr){2-3}
\cmidrule(lr){4-5}
\cmidrule(lr){6-7}
\cmidrule(lr){8-9}
& AUC & F1-Macro
& AUC & F1-Macro
& AUC & F1-Macro
& AUC & F1-Macro \\
\midrule
0.00 
& 94.83 ± 0.67 & 88.52 ± 0.51 
& 89.76 ± 0.98 & 78.41 ± 0.56 
& 94.15 ± 0.65 & 87.89 ± 0.42 
& 96.08 ± 0.45 & 91.31 ± 0.78 \\
0.25 
& \third{96.34 ± 0.59} & \third{90.15 ± 0.43} 
& \third{91.52 ± 0.91} & \third{80.33 ± 0.45} 
& \third{95.84 ± 0.53} & \third{89.47 ± 0.34} 
& \third{97.65 ± 0.32} & \third{92.84 ± 0.68} \\
0.50 
& \best{97.41 ± 0.54} & \best{91.17 ± 0.35} 
& \best{92.68 ± 0.84} & \best{81.55 ± 0.37} 
& \best{96.79 ± 0.48} & \best{90.42 ± 0.26} 
& \best{98.37 ± 0.25} & \best{93.72 ± 0.61} \\
0.75 
& \second{96.82 ± 0.61} & \second{90.64 ± 0.42} 
& \second{92.05 ± 0.89} & \second{80.92 ± 0.42} 
& \second{96.31 ± 0.51} & \second{89.86 ± 0.31} 
& \second{97.94 ± 0.29} & \second{93.25 ± 0.64} \\
1.00 
& 95.12 ± 0.72 & 89.23 ± 0.56 
& 90.43 ± 1.05 & 79.18 ± 0.53 
& 94.92 ± 0.62 & 88.54 ± 0.39 
& 96.81 ± 0.38 & 92.14 ± 0.73 \\
\bottomrule
\end{tabular}
}
\label{app:sens_alpha_full}
\end{table*}

\section{Baselines}
\label{app:baseline}
We conduct a comprehensive comparison of FADE with seventeen baseline algorithms, which fall into two categories: generic GNN models and graph anomaly detection models.
The generic GNN models include MLP \cite{MLP}, GCN \cite{GCN}, GraphSAGE \cite{GraphSAGE}, GAT \cite{GAT}, GIN \cite{GIN}, and GATv2 \cite{GATv2}.
The details of these graph anomaly detection models are introduced as follows:
\begin{itemize} 
	\item \textbf{CARE-GNN \cite{CARE-GNN}} proposes a camouflage-resistant method to improve aggregation process using three distinct units specifically designed to address camouflage challenges.    
 
 	\item \textbf{GraphConsis \cite{GraphConsis}} is a heterogeneous graph neural network designed to address three types of inconsistencies in the graph anomaly detection task: context, feature, and relation inconsistencies.

 	\item \textbf{PC-GNN \cite{PC-GNN}} is a specifically designed method for imbalanced learning.
    It selects nodes and edges with a label-balanced sampler, addressing the class imbalance problem.

 	\item \textbf{BWGNN \cite{BWGNN}} is a spectrum-based method to analyze graph anomalies.
    It employs Beta graph wavelets to create localized band-pass filters in the spectral and spatial domains for anomaly detection.

 	\item \textbf{H2-Fdetector \cite{H2}} is a graph anomaly detector with homophilic and heterophilic interactions.
    It adopts a distinct aggregation strategy for detected heterophilic edges, which is trained with an additional loss function.
    
 	\item \textbf{GHRN \cite{GHRN}} proposes a method based on graph theory to eliminate heterophilic edges and utilize the new graph for model training, which adopts BWGNN as its backbone.
  
   	\item \textbf{GDN \cite{GDN}} divides the node representation into class features and surrounding features. 
    It ensures that nodes in the same class have similar class features through class constraint, while neighboring nodes have similar surrounding features through connectivity constraint.
    
   	\item \textbf{GAGA \cite{GAGA}} is designed for anomaly detection in multi-relation graphs based transformer.
    It addresses the problems of low homophily and low label utilization in graph anoamly detection through population aggregation and enhancement of learnable encoding.

   	\item \textbf{DiG-In-GNN \cite{DiG}} utilizes reinforcement learning (RL) to select node-wise neighbor. 
    The RL-based neighbor selector helps facilitate message passing by aggregating relevant neighbors, thereby enhancing the detection performance with new node representations.
    
   	\item \textbf{ConsisGAD \cite{consisGAD}} designs a novel learnable data augmentation mechanism to utilize large amounts of unlabeled data for consistency training to improve anomaly detection performance.
    
   	\item \textbf{SpaceGNN \cite{SpaceGNN}} thoroughly investigates the benefits of incorporating multi-spatial information for anomaly detection tasks from both empirical and theoretical perspectives.

    \item \textbf{TAQ-GAD \cite{TAQ-GAD}} is a semi-supervised graph anomaly detection framework that quantifies node abnormality using two topological metrics: Node Boundary Score (NBS) and Node Isolation Score (NIS), which guides pseudo-anomaly selection and refines them through risk-based label flipping and virtual anomaly centres.

\end{itemize}

\section{Additional Results}
\subsection{LLM Variants}
\label{app:llm_var}
Table \ref{app:llm_variants} reports the performance of SignGAD with different LLM variants. 
Overall, stronger LLMs consistently lead to better performance across all datasets and metrics, indicating that the quality of workflow planning plays an important role in few-shot graph anomaly detection. 
GPT-4.1 achieves the best results in all evaluation cases, while GPT-3.5 consistently ranks second, showing that more capable planning agents can construct more suitable detection workflows. 
Compared with LLaMA3-8B, GPT-4.1 improves AUC by 1.73\%, 1.86\%, 1.74\%, and 1.72\% on Amazon, YelpChi, T-Finance, and T-Social, respectively, with similar gains on F1-Macro. 
These results suggest that LLM-based planning affects not only the ranking quality of anomaly scores but also the calibrated binary predictions. 
Nevertheless, all LLM variants achieve competitive performance, demonstrating that the proposed workflow design is not tied to a specific LLM backbone. 
By using stronger planning agents to construct task-conditioned workflows, SignGAD better organizes graph encodings, anomaly evidence, and detector designs, verifying the importance of agentic workflow planning in few-shot graph anomaly detection.

\subsection{Varying Levels of Supervision}
\label{app:vary}
Table~\ref{app:train_ratio_full} reports the complete results under different training ratios. 
Overall, SignGAD achieves the best AUC across all dataset-ratio settings and obtains the best F1-Macro in most cases, showing its stable effectiveness under different supervision levels. 
The advantage is especially clear on YelpChi and T-Social, where SignGAD consistently outperforms strong baselines by large margins, indicating that task-conditioned workflows can better organize anomaly evidence when graph tasks are challenging. 
On T-Finance, SignGAD maintains the best AUC across all ratios, while its F1-Macro becomes the best when the training ratio increases to 5\% and 10\%, suggesting that its ranking quality can be effectively converted into stronger binary predictions with more supervision. 
These results further verify that workflow-level customization improves both few-shot robustness and scalability under increased supervision.

\subsection{Ablation Study}
\label{app:ablation}
We construct five ablation variants to evaluate the contribution of each component. 
\textit{w/o LLM} replaces LLM-based planning with a fixed rule-based workflow template, removing the ability to design workflows from textual task descriptions and graph statistics. 
\textit{w/o Evid.} discards the evidence functions selected by the workflow, so the detector only receives graph-context representations. 
\textit{w/o Bank} replaces the workflow detector bank with a single default detector, eliminating detector-agent selection. 
\textit{w/o Search} skips validation-based workflow selection and directly uses the default candidate workflow without calibration. 
\textit{w/o Refit} disables guarded final refit and keeps the detector obtained after workflow search.

Table \ref{app:ablation_full} presents the full ablation results under the 1\% training ratio. 
Removing any component leads to consistent performance degradation across all datasets, verifying that each module contributes to the final detection workflow. 
Among all variants, \textit{w/o Refit} causes the largest performance drop in most cases, showing that guarded final refit is important for refining the selected workflow under few-shot supervision. 
The degradation of \textit{w/o Search} demonstrates the necessity of validation-based workflow calibration, since directly using the default workflow weakens both ranking quality and binary classification performance. 
Compared with \textit{w/o Evid.}, the full SignGAD achieves clear improvements on all datasets, indicating that explicitly organized anomaly evidence provides useful task-specific signals beyond graph-context representations. 
The consistent drop of \textit{w/o LLM} further confirms that LLM-based planning helps construct more suitable workflows from task information. 
Finally, the performance gap between \textit{w/o Bank} and SignGAD shows that detector-agent selection contributes to matching different graph tasks with suitable inductive biases. 
Overall, these results verify that SignGAD benefits from the joint design of agentic workflow planning, contextual evidence construction, detector selection, validation calibration, and guarded refinement.

\subsection{Evidence Function Number Analysis}
\label{app:function}
Table \ref{app:sens_evidence_num_full} reports the sensitivity analysis of the evidence function number $K$. 
When $K=0$, SignGAD removes explicit evidence functions and only relies on graph-context representations, leading to the weakest performance across all datasets. 
As $K$ increases, both AUC and F1-Macro generally improve, showing that incorporating more anomaly evidence helps the detector access task-relevant signals beyond latent node representations. 
The best performance is consistently achieved at $K=4$, indicating that a compact set of evidence functions is sufficient to capture useful contextual and structural anomaly patterns. 
When $K$ further increases to 5 or 6, the performance slightly decreases, suggesting that excessive evidence functions may introduce redundant or noisy signals. 
These results demonstrate that explicit evidence construction is beneficial for SignGAD, while selecting an appropriate evidence budget is important for stable few-shot graph anomaly detection.

\subsection{Candidate Workflow Number Analysis}
\label{app:candidate}
Table \ref{app:sens_workflow_num_full} reports the sensitivity analysis of the candidate workflow number $N_{\omega}$. 
When $N_{\omega}=5$, the performance is relatively lower, indicating that a small workflow pool may not provide enough candidate designs for validation-based selection. 
Increasing $N_{\omega}$ from 5 to 20 consistently improves both AUC and F1-Macro across all datasets, showing that a richer workflow space helps SignGAD discover more suitable graph encodings, anomaly evidence, and detector designs. 
The performance becomes saturated when $N_{\omega}$ further increases to 30, where the results are very close to those obtained with $N_{\omega}=20$. 
This suggests that overly large workflow pools provide limited additional benefit while increasing the search cost. 
Therefore, we set $N_{\omega}=20$ as the default setting, which achieves a strong trade-off between detection performance and workflow search efficiency.

\subsection{Validation Balance Coefficient Analysis}
\label{app:validation}
Table \ref{app:sens_alpha_full} reports the sensitivity analysis of the validation balance coefficient $\alpha$. 
When $\alpha=0$, workflow selection mainly emphasizes binary classification quality, while $\alpha=1$ focuses more on ranking quality. 
Both extreme settings lead to lower performance, indicating that relying on either objective alone is insufficient for selecting reliable workflows. 
The best results are consistently achieved at $\alpha=0.50$ across all datasets and metrics, showing that balancing ranking quality and binary prediction quality is important for validation workflow search. 
When $\alpha$ increases to $0.75$, the performance remains competitive but slightly lower than $\alpha=0.50$, suggesting that excessive emphasis on ranking scores may weaken calibrated binary decisions. 
Therefore, we set $\alpha=0.50$ as the default value in SignGAD.

\end{document}